\pdfoutput=1
\documentclass{article}
\usepackage{iclr2027_conference,times}

\usepackage{graphicx}
\usepackage{amsmath}
\usepackage{amssymb} 
\usepackage{booktabs}
\usepackage{xcolor}
\usepackage{url}  
\usepackage{physics}
\usepackage{hyperref} 
\usepackage{mathtools}
\usepackage{microtype}

\newcommand{\Nres}{N_{\rm res}}
\newcommand{\ASRNM}{{\rm ASR@}(N,M)}
\newcommand{\AFRNM}{{\rm AFR@}(N,M)}

\title{Escaping Alignment: A Physical Trap Model of Best-of-$N$ Jailbreaking}

\iclrfinalcopy
\author{Marco Biroli \\
Physical Science Division \\
Chicago, IL 60637, USA \\
\texttt{mbiroli@uchicago.edu}}

\begin{document}

\maketitle
\lhead{Preprint. Under review.}

\begin{abstract}
Best-of-$N$ jailbreaking (BoN) bypasses safeguards of aligned models by drawing $N$ independent augmentations of an unsafe prompt and sampling $M$ completions of each. Previous works have shown that the attack success rate (ASR) seems to follow a power-law in $N$, which we challenge. The exponent drifts with $N$, and the reported exponential crossover is consistent with a finite-size effect of the adversarial dataset. Little work has been done to explore the entire two-budget ($N, M$) attack surface as well as its dependence on the generation temperature $T$. We introduce a simple barrier model where each prompt has a baseline safety level and each augmentation a random thermally activated barrier. Then five numbers, each backed by an interpretable safety mechanism, determine the entire ($N, M$) attack surface. They extrapolate predictions from $N \leq 100$ to $N = 10^4$, collapse five distinct models on the same scaling function and predict ASR at different temperatures from the one they were fitted at. The extrapolation holds under two automated judges and on held-out prompts.
\end{abstract}

\begin{figure}[h]\centering
  \begin{minipage}[c]{0.65\linewidth}\centering\includegraphics[width=\linewidth]{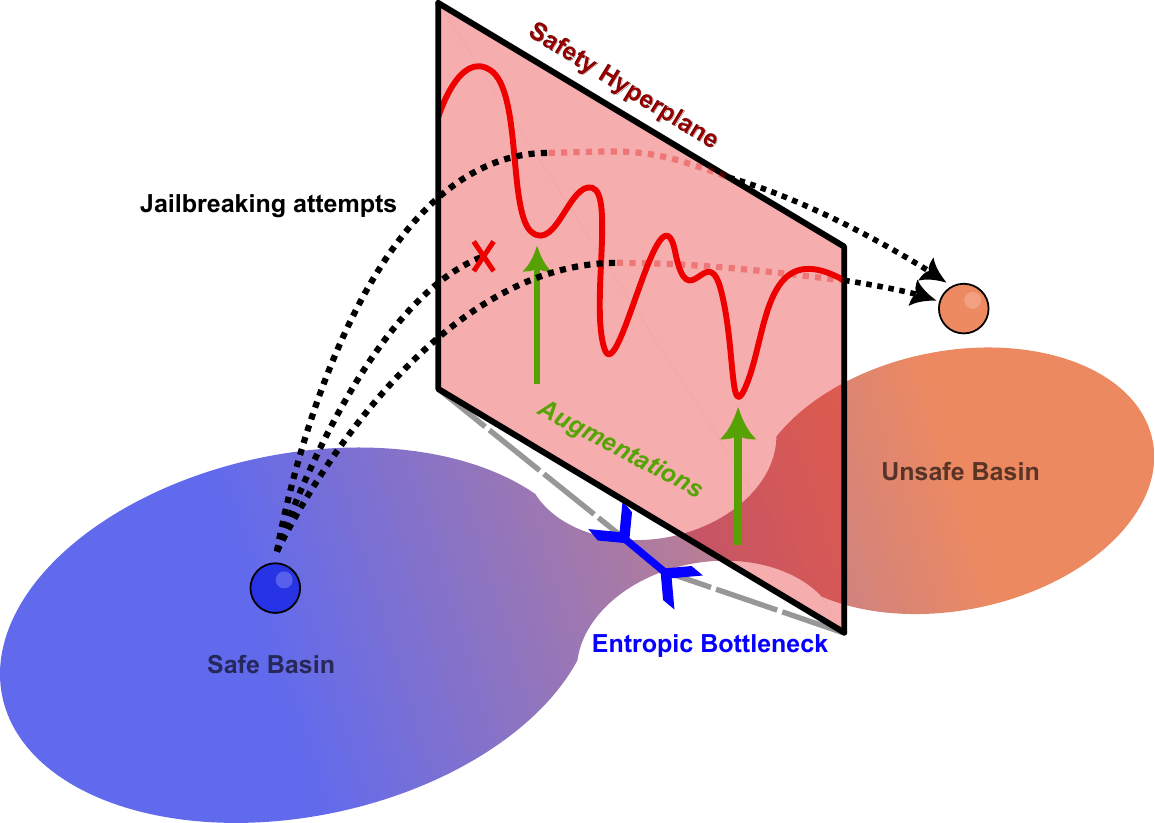}\end{minipage}\hfill
  \begin{minipage}[c]{0.32\linewidth}\caption{\label{fig:model} \textbf{The mechanism.} Generation starts in the safe
basin (blue) and a jailbreak means reaching the unsafe one (orange). The two
communicate only through an \emph{entropic bottleneck} (blue arrow). At the bottleneck lies a safety hyperplane (red), with energetic barriers preventing thermal crossing from the safe to unsafe basin. Augmentations select saddles of this energetic potential. Each completion (dotted) is an independent Arrhenius attempt over
that \emph{fixed} saddle.}\end{minipage}
\end{figure}

\section{Introduction}

Best-of-$N$ (BoN) jailbreaking~\citep{hughes2024bon} defeats aligned language models by repeatedly perturbing a refused prompt and sampling completions. Its attack success rate grows as
a power law in the number of attempts~\citep{hughes2024bon,feng2026saber}, sometimes reported as crossing over to exponential growth~\citep{crossover2026}. The exponent and the crossover budget are the quantities a defender cares about since they give the best estimate of the resistance of the model to adversarial efforts. These values are usually obtained by fitting curves to a benchmark of adversarial attacks. However, this does not allow us to understand \emph{why} an exponent takes the value it does and \emph{where} the failures stem from. Furthermore, it conflates several distinct sources of randomness. This paper takes an alternative route. We posit a physically inspired microscopic law, show that it reproduces the entire phenomenology of BoN attacks, and when fitted to experimental data is predictive for the whole two-budget surface. 

The physical model is schematically depicted in Fig.~\ref{fig:model}, and we are able to recover \emph{a posteriori} several of its elements directly from the residual stream.
Large Language Models (LLMs) transform discrete input tokens into a succession of continuous high-dimensional vectors called residuals which are the output of each transformer block. This residual stream has been shown to behave surprisingly linearly \citep{park2023linear}. It is therefore a good candidate to ground our physical model. Each sequence of discrete text $b$ is mapped to the residual of its last token $r_b \in \mathbb{R}^{d_{\rm model}}$ at layer 30 of 36 in Qwen2.5-3B-Instruct (App.~\ref{sec:exp}). We will now postulate three behaviors in this approximately linear residual space. Firstly, safety is a one-dimensional concept and therefore a single hyperplane will separate safe from unsafe text \citep{arditi2024refusal,wollschlager2025refusalcones}. Secondly, generation at temperature $T$ can be seen as simple thermally activated dynamics in the residual space. Finally, we can define an energy landscape in this hyperplane which will bound the thermal dynamics. As we will see in Section~\ref{sec:model}, this predicts that the probability that a prompt $b$ with augmentation $a$ yields an unsafe completion can be written as 
\begin{equation} \label{eq:p-def}
  p_{b, a} = \min\left(1, \; e^{-\alpha_b - \eta_{b, a} / T}\right) \;,
\end{equation}
where $\alpha_b$ denotes a Gaussian entropic bottleneck and $\eta_{b, a}$ exponentially distributed energetic saddles depicted in Fig.~\ref{fig:model}. Here $T$ is the generation temperature. Across prompts, $\alpha_b$ has mean $\mu_\alpha$ (the baseline safety) and spread $\sigma_\alpha$. The energetic saddles of prompt $b$ have exponential rate $\beta_b$, with $\log\beta_b$ of mean $\mu_\beta$ (the susceptibility to augmentations) and spread $\sigma_\beta$. The entropic ($\alpha$) and energetic $(\beta$) barriers are correlated via $\rho$. These are the five parameters of the model (notation in App.~\ref{app:notation}). We generalize the attack success rate (ASR) and attack failure rate (AFR) with $N$ augmentations, ${\rm ASR@}N \equiv 1 - {\rm AFR@}N$, to a Best-of-$N$/$M$ attack (BoNM) with $M$ completions per augmentation. The success rate $\ASRNM \equiv 1 - \AFRNM$ is the probability that at least one of the $NM$ completions is unsafe:
\begin{equation} \label{eq:ASRNM}
  \ASRNM = \left\langle 1 - \left[\left\langle (1 - p_{b,a})^{M} \right \rangle_{a \sim {\rm augmentations}}\right]^N \right\rangle_{b \sim {\rm unsafe~prompt~dataset}} \;.
\end{equation}
For $M = 1$ this reduces to the usual ${\rm ASR@}N \equiv {\rm ASR@}(N,1)$. Eq.~(\ref{eq:ASRNM}) can be understood as: at least one of the attacks succeeding is the complement of none of the $N$ augmentations succeeding and the probability of a given augmentation failing is that all $M$ completions fail which happens with probability $(1 - p_{b, a})^M$. Surprisingly, in direct contradiction with established literature \citep{hughes2024bon,feng2026saber} our model predicts that the ASR {\bf does not} follow a clean power-law. 

Section~\ref{sec:background} reviews related work, Section~\ref{sec:model} introduces the model and derives its predictions, Section~\ref{sec:validation} validates them experimentally and Section~\ref{sec:consequences} discusses what they entail. 

\section{Background and related work}
\label{sec:background}

\emph{Best-of-$N$ jailbreaking.} A strength of Best-of-$N$ jailbreaking
\citep{hughes2024bon} is its ability to attack black-box models without any information about what is happening underneath the hood. This has also led to most analyses of BoN jailbreaking to be model agnostic and restricted to fitting curves on a benchmark \citep{feng2026saber,crossover2026} from which two broad phenomenologies emerged. Firstly, the attack success rate seems to follow a power-law in $N$ \citep{hughes2024bon}. Secondly, at large enough $N$ the power-law crosses over into an exponential growth \citep{crossover2026}. To our knowledge, after \citet{hughes2024bon} ablations showing that multiple generations $M$ did not improve attack efficiency, no work has studied the full $(N, M)$ attack surface, nor its temperature dependence. In this paper we put forward a model for the entire $(N, M)$ attack surface and its explicit temperature dependence. This allows us to explain why most works seem to favor $M^\star = 1$. Surprisingly, we see that more compute-efficient attacks can be obtained for $M^\star > 1$. Furthermore, contrary to the reported literature we show that the attack success rate {\bf does not} follow a clean power-law, and that the exponential crossover is consistent with a finite-size effect of the datasets used to fit these attacks rather than a change of law. Most of our experiments use a white-box augmentation kernel, namely Gaussian noise on the input embeddings \citep{schwinn2024softprompt,geisler2024pgd, robey2023smoothllm,cohen2019smoothing}. App.~\ref{sec:kernel} shows that the structure of the model also holds for the character-level kernel of \citet{hughes2024bon}. We expect it to hold for any kernel applied independently and identically to every augmentation, but we have tested only these two. Therefore iterative attacks such as GCG \citep{gcg}, AutoDAN \citep{liu2024autodan}, or PAIR \citep{chao2023pair} are not described by our framework.

\emph{Power laws.} The fact that averaging stochastic individual success/failure rates lead to power laws (or slowly-varying power-laws) is well established
\citep{hutter2021learning,schaeffer2025powerlaws,levi2024,levi2026hardtail,kazdan2025}. Closest to us, \citet{feng2026saber} assume a Beta-distributed per-prompt rate, whose power-law left tail yields an exact power law in $N$, and extrapolate ASR@$N$ from a hundred samples to a thousand. A Beta distribution also appears in our model, but at a different level, and the two should not be confused. The Beta of \citet{feng2026saber} describes how the single-attempt rate $\langle p_{b,a}\rangle_a$ varies \emph{across prompts} $b$. Ours emerges from the microscopic model and describes how $p_{b,a}$ varies \emph{across augmentations} $a$ of one fixed prompt (Eq.~(\ref{eq:p-dist})). At $M=1$ this inner distribution enters only through its mean, so it plays no role in ASR@$N$ and only shapes the $M$ channel, where it yields a per-prompt power law in $M$. The $N$ channel is instead set by the across-prompt law, which in our model is log-normal ($E_b$ Gaussian, Eq.~(\ref{eq:def-E})) rather than Beta. The left tail of a log-normal decays faster than any power, which replaces the fixed exponent of \citet{feng2026saber} by a slowly varying one (Eq.~(\ref{eq:effective-exponent})). 

\emph{Glassy systems.} The model we put forward is inspired by canonical works in glassy systems. Trap models with exponentially distributed depths are the
classical description of activated dynamics in glasses \citep{bouchaud1992,derrida1981rem}. The combination of an entropic and an energetic barrier in Arrhenius escape is inspired by known physical models from reaction-rate theory \citep{hanggi1990reaction}. Recently, \citet{crossover2026} also put forward a glassy dynamical model to explain BoN jailbreaking. However, their model differs from ours in two aspects. First, they describe a single budget, whereas we resolve the two-budget $(N, M)$ surface and find that the two channels carry different exponents. Second, their model is based on interpreting generated tokens as individual degrees of freedom with an energy landscape defined in token space. 

\section{The microscopic model}
\label{sec:model}

Inspired by a physical mental model we postulate the following microscopic Ansatz
\begin{equation}
  \begin{gathered}
  p_{b, a} = \min(1, \; e^{-\alpha_b - \eta_{b, a}/T}) \mbox{~~where~~} \eta_{b, a} \sim {\rm Exp}[\beta_b] \mbox{~~and} \\
  {\begin{pmatrix} \alpha_b \\ \log \beta_b \end{pmatrix} \sim \mathcal{N}\!\left[\begin{pmatrix} \mu_\alpha \\ \mu_\beta \end{pmatrix}, \begin{pmatrix} \sigma_\alpha^2 & \rho\,\sigma_\alpha\sigma_\beta \\ \rho\,\sigma_\alpha\sigma_\beta & \sigma_\beta^2 \end{pmatrix}\right]} \;.
  \end{gathered} \label{eq:ansatz}
\end{equation}
Our model has only 5 free parameters $(\mu_\alpha, \sigma_\alpha, \mu_\beta, \sigma_\beta, \rho)$ to describe the entire $(N, M)$ attack surface. Readers who might not be interested in the physical mental picture behind Eq.~(\ref{eq:ansatz}) can jump to Sec.~\ref{sec:validation}.

The main assumption of the model is that the process of generating unsafe completions is governed by a high-dimensional energy landscape with two stable basins representing the safe and unsafe regions. However only two components of this landscape will come into play: the entropic bottleneck connecting the stable basins and the one-dimensional projection along a hyperplane separating the safe and unsafe basins. Motivated by existing experimental results we postulate the following structure:
\begin{itemize}
  \item {\bf Baseline safety.} Alignment training \citep{rafailov2023dpo, ouyang2022rlhf} finetunes the model to increase the log-odds of `safe' completions compared to `unsafe' completions. The alignment objectives in \citet{rafailov2023dpo, ouyang2022rlhf} create an entropic bottleneck between the two. We assume the model refuses unsafe completions of a prompt $b$ with a base entropic barrier $\alpha_b$. We further assume that these bottlenecks are Gaussianly distributed $\alpha \sim \mathcal{N}(\mu_\alpha, \sigma^2_{\alpha})$ over the dataset.
  \item {\bf Linearity.} Given that residual stream spaces have shown to be surprisingly linear \citep{park2023linear,zou2023repe}, we assume that a single hyperplane along the entropic bottleneck separates the safe from unsafe completions. 
  \item {\bf Augmentations select minima of exponential height.} We postulate that there exists an energy landscape within the hyperplane which separates the safe/unsafe completions. Given the usual glassiness of these systems we assume the landscape to be extremely rugged with exponentially distributed minima \citep{bouchaud1992,derrida1981rem}. Then each augmentation corresponds to choosing a height $\eta_{b, a} \sim {\rm Exp}[\beta_b]$ in this rugged landscape to try and cross-over to the other side. 
  \item {\bf Saddle heights are a property of the prompt.} We assume that the statistics of the saddle heights (which are governed by $\beta_b$) depend only on the prompt $b$. Hence, crucially, $\beta_b$ is kept constant for any augmentation $a$. We further postulate that $\beta_b$ is log-normally distributed $\log \beta \sim \mathcal{N}(\mu_\beta, \sigma^2_\beta)$, jointly Gaussian with $\alpha_b$ with correlation $\rho$; nothing in the mechanism requires the bottleneck of a prompt and the statistics of its saddles to be independent.
  \item {\bf Completions.} Once a prompt $b$, and augmentation $a$ are chosen they define an entropic bottleneck $\alpha_b$, a height $\eta_{b, a}$ and generation at temperature $T$ will cross over the saddle with Arrhenius probability $p_{b,a} = \min(1, e^{-\alpha_b-\eta_{b,a}/T})$ \citep{hanggi1990reaction}.
\end{itemize}
We introduce the {effective jailbreak rate at completion budget $M$} 
\begin{equation} \label{eq:Rb}
  R_b(M) = - \log \langle (1 - p_{b, a})^M \rangle_{\eta} \;.
\end{equation}
The quantity $R_b(M)$ represents: `given one augmentation of a fixed prompt $b$ completed $M$ times, what frequency of jailbreaks can we expect?'. A prompt survives an $(N, M)$ attack with probability
\begin{equation} \label{eq:survival-prob}
  Q_b(M)^N \equiv e^{-N R_b(M)} \;,
\end{equation}
where $Q_b(M)$ is the survival probability of a single augmentation. Then
\begin{equation} \label{eq:ASRNM-Rb}
  \ASRNM = 1 - \langle e^{- N R_b(M)} \rangle_b \;.
\end{equation}
Now expanding Eq.~(\ref{eq:Rb}) yields
\begin{equation} \label{eq:Rb-2F1}
  R_b(M) = -\log \left[\beta_b \int \dd \eta \; e^{-\beta_b \eta} (1 - e^{-\alpha_b - \eta / T})^M\right] = -\log [ \,_{2}F_1\left( -M, \beta_b T, 1 + \beta_b T, e^{-\alpha_b}  \right) ] \;,
\end{equation}
where $\,_2F_1(a, b, c, z)$ is the hypergeometric $\,_2F_1$ function \citep{dlmf}. 
Given that we are interested in the large-$N$ limit, the behavior will be dominated by the hardest prompts, i.e. $\alpha_b \gg 1$. Furthermore, the usual working regime for BoN attacks has $M \sim \mathcal{O}(1)$ and hence $M e^{-\alpha_b} \ll 1$ which allows us to simplify the hypergeometric function to 
\begin{equation} \label{eq:logRb}
  \log R_b(M) \simeq \log M - \alpha_b + \log(\beta_b T) - \log(1 + \beta_b T) \;.
\end{equation}
Neglecting the clamp of Eq.~(\ref{eq:p-def}) here and below changes ${\rm ASR@}N$ by at most $0.003$ ($0.012$ for Llama; App.~\ref{app:estimator}). Notice that averaging over all possible augmentations $a$ for a fixed prompt $b$ we can define an effective prompt barrier
\begin{equation} \label{eq:def-E}
  E_b = -\log\langle p_{b,a}\rangle_a = \alpha_b - \log(\beta_b T) + \log(1 + \beta_b T) \;,
\end{equation}
so that $\log R_b(M) \simeq \log M - E_b$. In other words, the jailbreaking frequency $R_b(M) \simeq M e^{-E_b}$ corresponds to $M$ parallel jumps over an effective barrier of height $E_b$. Given that $\alpha_b$ is Gaussian and $\beta_b$ is log-normal $E_b$ is well approximated by a Gaussian of mean $\mu_E$ and variance $\sigma^2_E$, which expanding $\log(1 + 1/\beta_b T)$ about $\log\beta_b T = m \equiv \mu_\beta + \log T$ gives in closed form:
\begin{equation} \label{eq:sigE}
  \mu_E \simeq \mu_\alpha + \log\left(1 + e^{-m}\right) + \tfrac{1}{2}\kappa(1 - \kappa)\,\sigma_\beta^2 \;, \qquad
  \sigma_E^2 \simeq \sigma_\alpha^2 + \kappa^2 \sigma_\beta^2 - 2 \rho\, \kappa\, \sigma_\alpha \sigma_\beta \;,
\end{equation}
where $\kappa = (1 + e^{m})^{-1}$. Against the exact moments under the fitted laws of Table~\ref{tab:params}, $\mu_E$ is accurate to $0.015$ for all five models and $\sigma_E$ to $0.03$, except for Llama-3.1-8B-Instruct, whose large $\sigma_\beta$ takes $\sigma_E$ beyond the second-order expansion ($2.01$ against $1.89$). Since $\log(1 + 1/\beta_b T)$ decreases with $\log\beta_b$, a positive $\rho$ can lower $\sigma_E$ below $\sigma_\alpha$, as can be seen for Qwen2.5-3B-Instruct. So Eq.~(\ref{eq:ASRNM-Rb}) becomes
\begin{equation} \label{eq:ASR-xi}
  \AFRNM  \equiv 1 - \ASRNM \simeq \left\langle \exp\left[ - e^{\log(N M) - \mu_E + \sigma_E \xi} \right]  \right\rangle_{\xi \sim \mathcal{N}(0, 1)} \;.
\end{equation}
The saddle point of Eq.~(\ref{eq:ASR-xi}) (see App.~\ref{app:saddle} for details) yields
\begin{equation} \label{eq:scaling-form}
  \AFRNM \simeq f_{\sigma_E}\left[ (N M) e^{-\mu_E}\right] \;,
\end{equation}
where
\begin{equation} \label{eq:f-def}
  f_\gamma(z) = \frac{1}{\sqrt{1 + W_0(\gamma^2 z)}} \exp\left[ - \frac{W_0(\gamma^2 z)^2 + 2 W_0(\gamma^2 z)}{2 \gamma^2} \right] \;,
\end{equation}
with $W_0$ being the Lambert function. A plot of this scaling function for various values of $\gamma$ is given in Fig.~\ref{fig:collapse}(c).  Writing $L \equiv \log(\gamma^2 z)$ and using $W_0(\gamma^2 z) \simeq L - \log L + \log L / L + \mathcal{O}(L^{-2})$, the asymptote of this scaling function when $z \to +\infty$ is 
\begin{equation} \label{eq:largez-full}
  f_\gamma(z) \simeq \frac{1}{\sqrt{1 + L - \log L}}\,
  \exp\left[ - \frac{\left(L - \log L\right)^2 + 2 L}{2 \gamma^2} \right] \;.
\end{equation}
Notice that the asymptote is not a clean power-law as is usually reported in the literature. In reality, it predicts a power law $f_\gamma(z) \propto z^{-\nu_{\rm eff}}$ with a slowly varying exponent
\begin{equation} \label{eq:effective-exponent}
  \nu_{\rm eff} \equiv -\dv{\log f_\gamma(z)}{\log z} = \frac{W_0(\gamma^2 z)}{\gamma^2} + \frac{W_0(\gamma^2 z)}{2\left[1 + W_0(\gamma^2 z)\right]^2} \stackrel{z \to +\infty}{\simeq} \frac{1}{\gamma^2} \log z \;.
\end{equation} 
Although less interesting, the low-budget (i.e. small $N$) asymptote can also be obtained in closed form. However, to do so we cannot take the small-$z$ limit of Eq.~(\ref{eq:f-def}) since $f_\gamma(z)$ was already the result of a large-$N$ saddle-point approximation. Instead we have to expand Eq.~(\ref{eq:ASR-xi}) directly for small $N$ which yields
\begin{equation} \label{eq:jensen}
  \AFRNM \simeq 1 - (N M)\, e^{-\mu_E + \sigma_E^2/2} \equiv 1 - z e^{\gamma^2/2}  \;.
\end{equation}

\section{Experimental validation}
\label{sec:validation}

\begin{table}[h]\centering
  \caption{The five parameters of each model, at $T = 1$, fitted by the marginal likelihood of Eq.~(\ref{eq:hier}) over all prompts; never-jailbroken prompts enter as censored observations. $\mu_E$ and $\sigma_E$ are the moments of the effective barrier $E_b$ of Eq.~(\ref{eq:def-E}) under the fitted law. Uncertainties are $\pm 1$ s.d.\ from a bootstrap over prompts ($200$ resamples). The amplitude of the BoN kernel noise is chosen such that the median jailbreak rate is in 1--5\%.}\label{tab:params}
\resizebox{\linewidth}{!}{%
\begin{tabular}{lccccccc}
\toprule
 & $\mu_\alpha$ & $\sigma_\alpha$ & $\mu_\beta$ & $\sigma_\beta$ & $\rho$ & $\mu_E$ & $\sigma_E$ \\
Model & Base alignment & Alignment fluctuations & Attack susceptibility & Susceptibility fluctuations & Correlation & Effective barrier & Barrier fluctuations \\
\midrule
Qwen2.5-7B (base)     & $0.56_{\pm 0.03}$ & $0.25_{\pm 0.03}$ & $+2.03_{\pm 0.09}$ & $0.84_{\pm 0.05}$ & $-0.33_{\pm 0.13}$ & $0.72_{\pm 0.03}$ & $0.32_{\pm 0.03}$ \\
Llama-3.1-8B-Instruct & $0.47_{\pm 0.15}$ & $0.72_{\pm 0.16}$ & $-1.97_{\pm 0.18}$ & $1.76_{\pm 0.13}$ & $-0.52_{\pm 0.14}$ & $2.75_{\pm 0.22}$ & $1.89_{\pm 0.15}$ \\
Qwen2.5-3B-Instruct   & $1.81_{\pm 0.03}$ & $0.63_{\pm 0.03}$ & $-0.90_{\pm 0.03}$ & $0.49_{\pm 0.02}$ & $+0.48_{\pm 0.04}$ & $3.07_{\pm 0.03}$ & $0.55_{\pm 0.02}$ \\
Qwen2.5-7B-Instruct   & $2.14_{\pm 0.08}$ & $1.29_{\pm 0.06}$ & ${\bf -2.09}_{\pm 0.06}$ & $1.11_{\pm 0.05}$ & $-0.22_{\pm 0.06}$ & ${\bf 4.40}_{\pm 0.10}$ & $1.76_{\pm 0.07}$ \\
Gemma-2-9B-it         & ${\bf 2.68}_{\pm 0.07}$ & $0.63_{\pm 0.06}$ & $+0.91_{\pm 0.10}$ & $0.36_{\pm 0.09}$ & $-0.46_{\pm 0.28}$ & $3.03_{\pm 0.07}$ & $0.69_{\pm 0.06}$ \\
\bottomrule
\end{tabular}}
\end{table}

\begin{figure}[t]\centering
\includegraphics[width=0.8\linewidth]{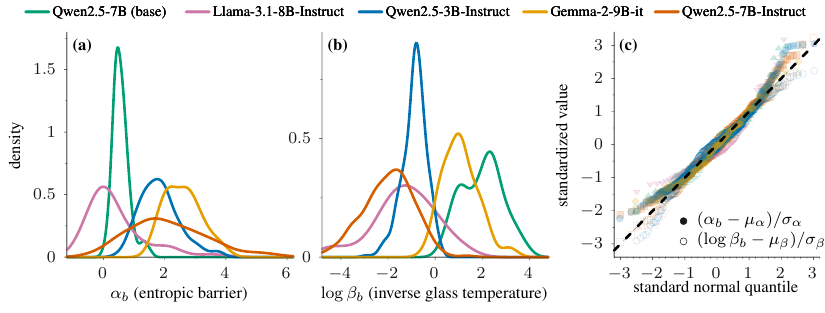}
\caption{\label{fig:params} {\bf Shared laws.} Across prompts, $\alpha_b$ is Gaussian and $\beta_b$ log-normal. Panel (c) shows how
standardized per-prompt parameters collapse onto $\mathcal N(0,1)$.}
\vspace{-6pt}
\end{figure}

We now show experimental evidence for this microscopic model in three families of open-source models with 3B to 9B parameters: Qwen \citep{qwen2025}, Llama \citep{llama3} and Gemma \citep{gemma2}. For each model we run a BoNM jailbreak on the first $400$ (Qwen-Instruct models) harmful prompts of AdvBench \citep{gcg} (see App.~\ref{sec:exp} for details). For each prompt we draw $N$ augmentations (Gaussian noise of variance $\sigma$ in embedding space or character-level noise) and for each augmentation we draw $M$ completions at temperature $T$. To have comparable results across models we choose $\sigma$ such that the average jailbreak rate is 1--5\% for each model. Most of our results are derived from a $N = 100$, $M = 300$ grid, then to probe the far-$N$ regime we further run an independent attack with $N = 10^4$ augmentations and $M = 1$ completion per prompt. Each completion is scored by an external judge (Granite-Guardian-3.0-2B \citep{padhi2024granite} and HarmBench-13B \citep{mazeika2024harmbench} as a control; every analysis on the two Qwen-Instruct models is repeated with HarmBench labels and with labels that require both judges to agree, App.~\ref{sec:judges}) and deemed unsafe if its harm probability exceeds 0.5. Leading to the empirical measure
\begin{equation}
  \hat p_{b, a} = \frac{1}{M} \sum_{i = 1}^{M} \vb{1}\{{\rm completion~} i {\rm ~of~prompt~} b {\rm ~with~augmentation~}a{\rm~is~jailbroken.}\} \equiv \frac{\hat k_{b,a}}{M}\;.
\end{equation}
Given a single prompt $b$, we can compute the individual jailbreak rate $\hat p_{b,a}$ and then average over the $N \gg 1$ augmentations $\langle \hat p_{b, a} \rangle_a$, our theory predicts that $-\log \langle p_{b, a} \rangle$ should be equal to
\begin{equation} \label{eq:log-avgp}
  -\log \langle p_{b, a} \rangle_a = -\log \langle e^{-\alpha_b - \eta/T} \rangle_a = \alpha_b - \log\left[\frac{\beta_b T}{1 + \beta_b T}\right] \;.
\end{equation}
Now instead of looking at the mean we can study the full distribution of $p_{b, a}$. Making the change of variables from $\eta$ to $p_{b,a}$ yields
\begin{equation} \label{eq:p-dist}
  {\rm Prob.}[p_{b,a} \, | \, b] = \beta_b T e^{\beta_b T \alpha_b } p^{\beta_b T - 1}
\end{equation}
Hence $\beta_b$ is clearly defined by the $p \to 0^{+}$ limit of this distribution and the upper support of this distribution is $e^{-\alpha}$. However, using the support edge as an estimator is extremely noisy. Instead we fit a soft-wall by Maximum Likelihood Estimation (MLE) using a Beta distribution $\hat p \sim B(\hat \beta_b T, \hat c_b)$ to model $\hat k_{b, a} \sim {\rm Binomial}(M, p)$. Then
\begin{equation}
  \log \langle \hat p \rangle = \log \frac{\hat \beta_b T}{\hat \beta_b T + \hat c_b}
\end{equation}
and equating with Eq.~(\ref{eq:log-avgp}) we get
\begin{equation} \label{eq:alpha-estimator}
  \hat \alpha_b = \log\left[ \frac{\hat \beta_b T + \hat c_b}{1 + \hat \beta_b T} \right] \;.
\end{equation}
Given Eq.~(\ref{eq:alpha-estimator}) and with an MLE fit we are now able to recover $\alpha_b$ and $\beta_b$ from numerical data. 
The goodness of fit of the MLE (over 400 prompts typical error $0.17$ in $\log Q_b(M)$ with $M \leq 300$) supports the Beta soft wall as a description of the within-prompt law on this dataset.
In Fig.~\ref{fig:params} we can see that the measured $\hat \alpha_b$ and $\log \hat \beta_b$ are well approximated by a bivariate Gaussian distribution. The raw distributions of $\hat \alpha_b$ and $\log \hat \beta_b$ of the two instruct models are very different but once centered and standardized they cannot be distinguished. The normal and log-normal \emph{families} are shared by all the models we tested, only their moments are model-specific.
Taking moments of these per-prompt estimates would, however, drop the prompts too hard to fit individually (fewer than three jailbreaks) and inflate the spreads with per-prompt estimation noise. We therefore use the per-prompt fits only to establish the form of the law, and estimate its five parameters $\theta$ from all prompts at once by maximizing the marginal likelihood
\begin{equation} \label{eq:hier}
  L(\theta) = \prod_b \int \dd\alpha_b \, \dd\log\beta_b \; \mathcal{N}(\alpha_b, \log\beta_b \,|\, \theta) \prod_a {\rm BetaBin}\!\left(k_{b,a} \,\middle|\, M, \beta_b T, c_b\right) ,
\end{equation}
with $c_b$ tied to $\alpha_b$ by Eq.~(\ref{eq:alpha-estimator}). Hence, a never-jailbroken prompt is counted as a censored observation instead of being dropped (App.~\ref{app:estimator}). Table~\ref{tab:params} shows the fitted constants for every model.

\paragraph{Safety Results.} We can see in Table~\ref{tab:params} that the base model has essentially no barrier. Its median augmentation is jailbroken roughly half of the time. The instruct-tuned version of the same model raises $\mu_\alpha$ by a factor of $3.8$ and $\sigma_\alpha$ by a factor of $5.2$. The entropic barrier is therefore not a property of the architecture but a consequence of training/finetuning. Alignment also improves resistance to BoNM attacks by moving $\mu_\beta$ from $+2.0$ to $-0.9$ or $-2.1$. Between the two Qwen-Instruct models, the 7B model is both safer ($\mu_\alpha^{(7B)} > \mu_\alpha^{(3B)}$) and less susceptible to the BoNM attack ($\mu_\beta^{(7B)} < \mu_\beta^{(3B)}$). Interestingly not all models defend in the same way. Gemma favors an `entropic defense' (the $\alpha$ channel) whereas Llama is entirely aligned via an `energetic defense' (the $\beta$ channel).

\begin{figure}[t]\centering
\includegraphics[width=0.8\linewidth]{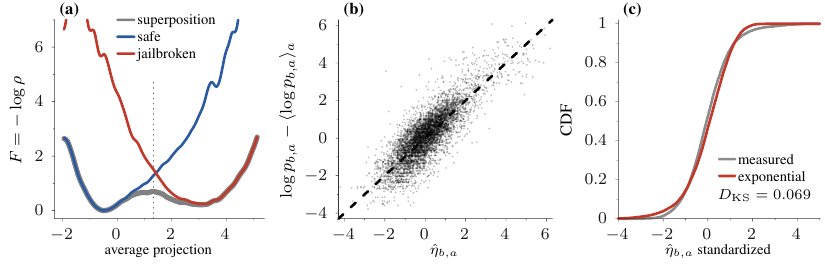}
\caption{\label{fig:physics} {\bf Geometry of the residual stream. } (a) Effective free energy along a supervised safety readout after balancing each population. Average projection along a single direction separates completions into two safe/unsafe regions. (b) The saddle $\eta_{b,a}$ predicted from the
augmented prompt alone against the measured per-augmentation
rate. (c) The distribution of the recovered
$\eta_{b,a}$, against the exponential assumption pushed through the same estimator.}
\vspace{-6pt}
\end{figure}

\paragraph{The physical picture.} To test the physical picture we now look for its constituting elements directly in the residual stream. For each prompt $b$ and augmentation $a$ we average the residuals of the completions the judge calls unsafe, subtract the average over those it calls safe, normalize, and then average the resulting unit contrasts over all $b$ and $a$ to compute a single `safety vector' $\hat{c}_{\rm hyper}$. The vector $\hat c_{\rm hyper}$ is the normal of the safety hyperplane of Fig.~\ref{fig:model}, and a linear readout along it separates safe from unsafe completions well. %
Taking the average projection of each completion on the safety direction, the two safe/unsafe basins appear clearly in Panel (a) of Fig.~\ref{fig:physics}.

A second crucial assumption of our model is that the saddle $\eta_{b, a}$ is a property of the augmentation $a$. Consequently, it should be readable from the augmented prompt residual directly, without the need to look at the completion. To check this we train a linear probe $\hat c_{\rm saddle}$ to predict $\log p_{b, a} - \langle \log p_{b, a} \rangle_a$ from the centered residual $r_{b, a} - \langle r_{b, a} \rangle_a$ of the last token of the augmented prompt. As shown in Panel (b) of Fig.~\ref{fig:physics} not only are we able to recover $\eta_{b, a}$ ($R^2 = 0.69$ on held out validation), but the spread of the projection along $\hat c_{\rm saddle}$ is inversely correlated to $\beta_b$ (Spearman $-0.59$). We can also use this projection to verify that the exponential assumption $\eta_{b, a} \sim {\rm Exp}[\beta_b]$ is reasonable, which is shown in Panel (c) of Fig.~\ref{fig:physics}. Since the probe is a noisy and censored estimate of $\eta_{b,a}$, the comparison is made by pushing the exponential law through the same estimator. This readout can be repeated at every layer for each model. The held-out $R^2$ is always zero at the embedding layer and peaks at $80-95$\% depth in all five models. Both directions also control the transition when we intervene on them (App.~\ref{app:interventions}).

\subsection{Reconstruction and extrapolation}

\begin{figure}[t]\centering 
\includegraphics[width=0.8\linewidth]{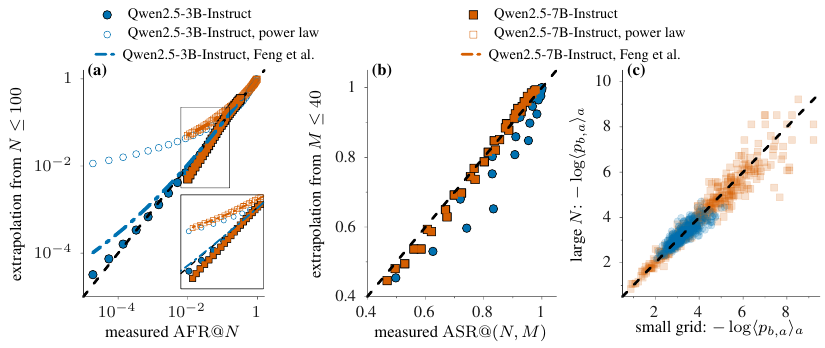}
\caption{\label{fig:trapval} {\bf Extrapolation (closer to the diagonal $=$ better).} We fit $(\alpha_b, \beta_b)$ on a small $N \leq 100$, $M \leq 40$ grid. Extrapolate its predictions and compare with two baselines: a power-law and \citet{feng2026saber} Beta fit, fitted on the same $N \le 100$ grid. In panel (a) we extrapolate to large $N \sim 10^{4}$ with the five-parameter law of Table~\ref{tab:params}, fitted on all $300$ completions; beyond the fit window the error in $\log_{10}$ survival is $0.09$ against $0.22$ for \citet{feng2026saber} (3B) and $0.09$ against $0.29$ (7B). The inset zooms into the boxed region, where the 7B predictions separate. In panel (b), we extrapolate to large $M \sim 300$. In panel (c), the effective barrier $-\log \langle p_{b,a} \rangle_a$ is measured two independent ways, on the $100 \times 300$ grid versus on the $10^4 \times 1$ grid. Both agree at $r = 0.85$ (3B) and $0.96$ (7B).} 
\vspace{-6pt}
\end{figure}

The first test is whether $(\mu_\alpha, \sigma_\alpha, \mu_\beta, \sigma_\beta{, \rho})$ are truly properties of only the model, i.e. independent of the prompt $b$ and the augmentation $a$. To check this we set $\hat c_b$ from Eq.~(\ref{eq:alpha-estimator}), and compute $\ASRNM$ from Eqs.~(\ref{eq:ASRNM-Rb}) on the $N, M \le 100$ grid. We compare this population-level fit with a separate Beta-distribution fit for each prompt. %
The five-number model matches the measured data with a Mean Absolute Error (MAE) of $0.036$. In comparison, the per-prompt fit leads to an MAE of $0.034$. In other words, the population model with only 5 fitting parameters is nearly as accurate as the per-prompt model with 800 parameters.

We now test whether this model is able to extrapolate the ASR beyond its fitting window. To do so, we fit the five-parameter family $(\mu_\alpha, \sigma_\alpha, \mu_\beta, \sigma_\beta{, \rho})$ on a small region of augmentations $N \leq 100$ and completions $M \leq 40$ (all $300$ for the extrapolation in $N$) and check how they perform at much larger values $N \sim 10^4$ and $M \sim 300$. Fig.~\ref{fig:trapval} shows how the small attack fit closely predicts large attack performances. In comparison, if we follow the usual practice in the literature \citep{hughes2024bon,feng2026saber} and fit a power-law to the $N \leq 100$ region and extrapolate to $N \sim 10^4$ the prediction would have been significantly off, as can be seen in Fig.~\ref{fig:trapval} panel (a). This is a consequence of the ASR not really following a power-law, as previously discussed in Eq.~(\ref{eq:effective-exponent}). Our law beats the Beta of \citet{feng2026saber} beyond the fit window on both models ($0.09$ against $0.22$ on Qwen2.5-3B-Instruct, $0.09$ against $0.29$ on Qwen2.5-7B-Instruct). On Qwen2.5-7B-Instruct the two separate within a few hundred augmentations (inset of Fig.~\ref{fig:trapval}(a)). On Qwen2.5-3B-Instruct they are equally accurate up to $N \approx 300$ and separate beyond it, where the survival falls below $10^{-3}$.

\paragraph{Other judges and unseen prompts.} Safety is subjective and judges disagree. While the notion of what is considered safe or unsafe might influence our fitted constants we find that our proposed mechanism remains a good predictor of the model's behaviour under all three label sets. On the same completions Granite flags $4.4\%$ as unsafe and HarmBench $0.3\%$, and a blind human audit finds that $14\%$ of Granite's flags and $79\%$ of HarmBench's align with our notion of harm (App.~\ref{sec:judges}). We therefore repeat the extrapolation of Fig.~\ref{fig:trapval}(a) with HarmBench labels and with labels that require both judges to agree. A stricter judge raises the fitted barriers, but the law extrapolates just as well: fitted at $N \le 100$, it predicts the survival up to $N = 10^4$ to within $0.02$--$0.09$ in $\log_{10}$ under all three label sets, against $0.10$--$0.29$ for \citet{feng2026saber} and $0.15$--$1.30$ for a power law (Table~\ref{tab:judgeextrap}). Under the stricter labels up to $43$ of the $400$ prompts are never jailbroken. The law also predicts prompts it was not fitted on: fitted on a random half of the prompts, it predicts the survival of the other half to within $0.02$--$0.07$ under the stricter labels, against $0.11$--$0.21$ for the Beta law (App.~\ref{app:heldout}).
\begin{figure}[t]\centering
{\includegraphics[width=0.8\linewidth]{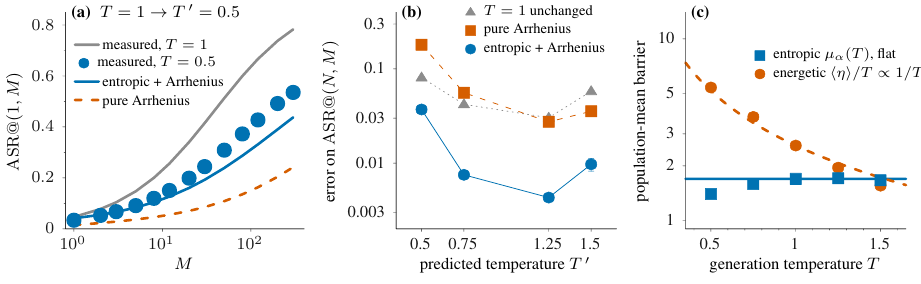}}
\caption{\label{fig:temp} {\bf Temperature dependence.} (a-b) Predicting a new decoding temperature from $T = 1$ data alone. Pure Arrhenius $p(T) = p(1)^{1/T}$ (dashed) overstates the effect of temperature in both directions. Our model keeps the entropic barrier outside the $1/T$ (Eq.~(\ref{eq:T-transfer})) with $\alpha_b$ fitted on the $T = 1$ data (blue) which reduces the error fourfold. (c) The barrier is entropic (constant in $T$), the saddles are energetic ($\propto 1/T$).}
\vspace{-6pt}
\end{figure}

{\bf Zero-shot temperature extrapolation.} Generation temperature $T$ also enters our model in a very specific manner: only the saddle crossings are thermal (via Arrhenius' law), neither the barrier $\alpha_b$ nor the inverse glass temperature $\beta_b = 1/T_g$ should move with $T$. Re-running the grid at five temperatures on $100$ prompts, with the augmentation seeds of the $T = 1$ grid, we find $\mu_\alpha$ and $\mu_\beta$ flat while the saddle height in units of $T$ follows $1/T$ (panel (c) of Fig.~\ref{fig:temp}). This is the direct evidence that $\alpha_b$ is entropic, i.e. independent of temperature whereas saddles are thermal. Furthermore, given the exact temperature structure, the whole ASR at a new temperature $T'$ follows from the $T = 1$ data through
\begin{equation} \label{eq:T-transfer}
  p_{b,a}(T') = e^{-\alpha_b (1 - 1/T')} \, p_{b,a}(1)^{1/T'} \;.
\end{equation}
Fig.~\ref{fig:temp} shows that Eq.~(\ref{eq:T-transfer}), with $\alpha_b$ fitted at $T = 1$, reduces the error by a factor of $3.7$ to $7.5$ relative to pure Arrhenius (i.e. $p_{b, a}(T) = e^{-(\alpha_b + \eta_{b, a})/T}$), at all four unseen temperatures.

\subsection{Scaling collapse}

\begin{figure}
  \centering
  \includegraphics[width=0.8\textwidth]{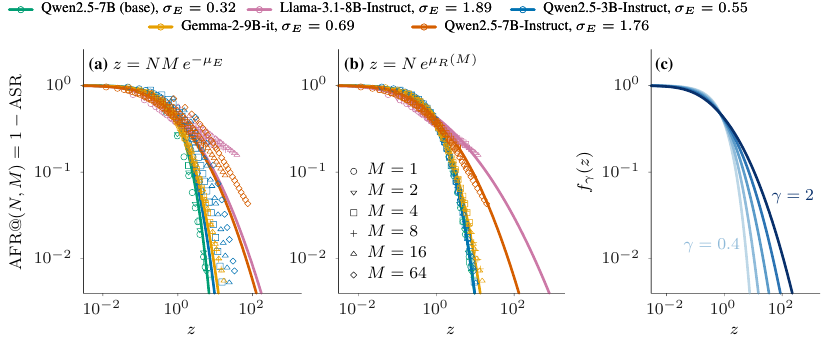}
  \caption{{\bf Scaling collapse of the AFR.} In panel (a) we show the small-$M$ closed-form leading order which although mostly accurate starts to deviate at large-$z$ for larger values of $M$. In panel (b) we use the $M$-agnostic form which remains accurate at large-$z$ for all $M$, with $\mu_R(M)$ and the width $\sigma_R$ read off the grid over all prompts (App.~\ref{app:saddle}). Finally in panel (c) we show a few examples of the scaling function $f_\gamma(z)$ for varying values of the shape parameter $\gamma$.}\label{fig:collapse}
\vspace{-6pt}
\end{figure}

The scaling form in Eq.~(\ref{eq:scaling-form}) makes a striking verifiable prediction. If we plot the AFR as a function of $z \equiv (N M)e^{-\mu_E}$ then all models should follow the same curve family $f_\gamma(z)$ defined in Eq.~(\ref{eq:f-def}). This collapse can be seen in Fig.~\ref{fig:collapse}. However, the exact scaling form in Eq.~(\ref{eq:scaling-form}) is only valid at $M \ll e^{\mu_\alpha}$ (which allowed us to simplify the hypergeometric function in Eq.~(\ref{eq:Rb-2F1})). Otherwise, the best we can do is keep the Gaussian approximation of $\log R_b(M) \sim \mathcal{N}(\mu_R(M), \sigma_R(M)^2)$ without knowing the closed form of $\mu_R(M)$ and $\sigma_R(M)^2$ which leads to panel (b) in Fig.~\ref{fig:collapse}. Both are read off the measured grid (App.~\ref{app:saddle}). We can see the clear separation between Qwen7B-Instruct and Llama ($\sigma_E \sim 1.8$) compared to Qwen3B-Instruct, Qwen7B-base and Gemma ($\sigma_E \sim 0.5$).

\subsection{Power-law and finite-size exponential crossover}
\label{sec:crossover-id}

\begin{figure}[t]\centering
{\includegraphics[width=0.8\linewidth]{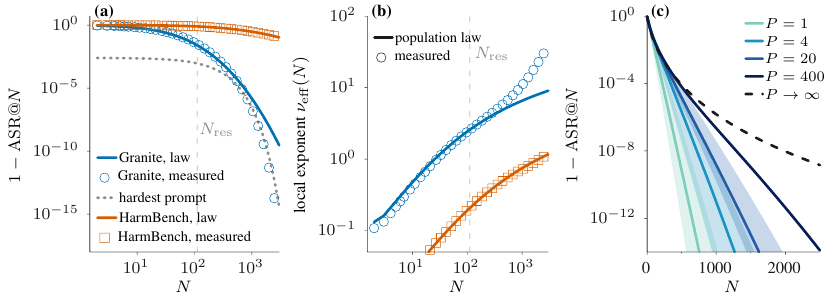}}
\caption{\label{fig:crossover} {\bf Fake power-law and finite-size exponential crossover.} For each prompt we draw $G = 10^4$ augmentations of which $H$ are jailbroken (unlike before all prompts are kept even those which are never jailbroken). The survival at budget
$N$ is estimated by $\binom{G-H}{N}/\binom{G}{N}$, i.e. that a random $N$-subset of the $G$ augmentations contains none of the $H$ jailbroken ones. (a) Measured population survival (points), the five-parameter population law from the grid fits
(solid), and the survival of the single hardest prompt of the dataset (dotted, Granite).
(b) The measured local exponent follows the predicted drift up to the resolution budget
$\Nres$ of the dataset, and then breaks away and `crosses
over' as the curve is taken over by the hardest prompt. (c) Survival of random subsets of $P$ prompts (Granite labels; median over $300$ subsets, band: interquartile range), against the population law ($P \to \infty$, dashed). Each subset follows the law until its own hardest prompt takes over and its survival becomes exponential. This break-away moves to larger $N$ as $P$ grows.}
\vspace{-6pt}
\end{figure}

We now turn to the usual power-law behavior reported in the literature \citep{hughes2024bon,feng2026saber}. We fit our model to the usual $N \leq 100, M \leq 300$ experimental grid and then run a separate attack on $N = 10^4$ augmentations (with $M = 1$) to probe the large-$N$ behavior. As shown in Fig.~\ref{fig:crossover} (blue, Granite labels) the five numbers of Table~\ref{tab:params} extrapolate to the large-$N$ quite well until $N \sim 10^{3}$ where the data seems to cross over to an exponential regime. In parallel, the measured local exponent $\nu_{\rm eff}(N)$ tracks the theoretical prediction and drifts from $0.1$ at $N = 2$ to above $2$ at $N \sim 10^2$--$10^3$ with no sign of a plateau, directly contradicting the usual power-law interpretations of BoN scaling. There is no fixed power-law exponent, only the running one of Eq.~(\ref{eq:effective-exponent}). We can clearly see in Fig.~\ref{fig:crossover}(a-b) that the exponents of both the Granite and stricter HarmBench judges are not constant at all. Instead they both increase as $\log N$.

Although \citet{crossover2026} have reported exponential corrections, we believe that this exponential crossover is a finite-size effect of the limited harmful dataset over which the ASR is computed. With a finite dataset of $P$ prompts the population average of Eq.~(\ref{eq:ASRNM-Rb}) is a sum of $P$ exponentials, and beyond the budget $\Nres = 1/\min_b \langle p_{b,a} \rangle_a$ at which even the hardest prompt of the dataset has been jailbroken, the survival is dominated by that single prompt and decays as a pure exponential $(1 - \langle p \rangle_a)^N/P$. As can be seen in Fig.~\ref{fig:crossover}, the measured curve follows the population law up to $\Nres$ and then breaks away. The exponent $\nu_{\rm eff}$ shoots up from $\sim 2$ to $\sim 10^2$. In our data this exponential behavior is a side-effect of the dataset size, not a property of the model. %
For the $400$-prompt AdvBench, $\Nres$ is approximately $10^2$ for Qwen-3B-Instruct for Granite, and beyond the $10^4$-augmentation arm for HarmBench, where the fitted law predicts $\approx 2 \times 10^4$. 

\section{Safety Consequences}
\label{sec:consequences}

{\bf Compute-optimal attack.}\label{sec:mstar} \looseness=-1 We can see from Eq.~(\ref{eq:scaling-form}) that $N$ and $M$ contribute equally to the ASR provided $M \ll e^{\mu_\alpha}$. When $M \sim e^{\mu_\alpha}$ more attempts yield diminishing returns. Hence, if completions are cheaper to sample than augmentations, an attacker should always use $M^\star \sim e^{\mu_\alpha}$ with an exact $M^\star$ whose closed form expression is a complicated implicit equation detailed in App.~\ref{app:mstar}. On the other hand, if augmentations are cheaper to sample than completions then there is no reason to go beyond $M^\star = 1$. 

{\bf What a defender should optimize.} BoN is a worst-case attack: Eq.~(\ref{eq:ASRNM-Rb}) is a Laplace transform of the barrier distribution and at any budget it is governed by the prompts of effective barrier $E_b \sim \log N$, i.e. extremely high barriers. A naive intuition would be that the best defense is obtained by a tight cluster of extremely high barriers, i.e. $\mu_E \gg 1$ and $\sigma_E \ll 1$. Surprisingly, Eq.~(\ref{eq:scaling-form}) shows that this is more nuanced. Fig.~\ref{fig:collapse} shows that higher values of $\sigma_E$ lead to safer models at large-$z$ but weaker models at smaller $z$. Since BoN is an extreme-value process, large $N$ eventually reaches any barrier height. However, by keeping $\sigma_E \gg 1$ we are `fighting' one extreme-value process with another. BoN will struggle to reach the few extreme barriers which are much higher than the mean, at the cost of sacrificing a few `easy to reach' low barriers.

{\bf Alignment protocols.} It is worth noting that different alignment schemes seem to lead to different defense types. Table~\ref{tab:params} shows that Gemma, Llama and Qwen have three different ways of defending against BoN. Gemma relies on a strong entropic bottleneck $\mu_\alpha$ but almost no energetic barrier $\mu_\beta$. In contrast, Llama has almost no entropic bottleneck but a large energetic barrier, and Qwen sits in between. Interestingly, Llama and Gemma used opposite alignment recipes: offline DPO \citep{rafailov2023dpo,llama3} versus on-policy RL \citep{gemma2}. How alignment training maps onto these five parameters is out of scope here but would be an interesting follow-up.

\section{Conclusion}
\label{sec:conclusion}

In this work, we put forward and validate experimentally a physics-inspired microscopic model for Best-of-$N$ jailbreaking and generalized Best-of-$N$ to Best-of-$NM$ by considering multiple completions for the same augmentation. We found that contrary to the usual literature the attack success rate does not follow a fixed power-law. We further showed that the reported exponential crossover is consistent with a finite-size effect of fitting jailbreaking behaviors on a finite adversarial dataset, set by its hardest prompts. Our theoretical model allows us to quantitatively compare the safety of different models as shown in Table~\ref{tab:params} and also makes predictions for best defensive practices to follow. 

\subsubsection*{Ethics statement}

\emph{Limitations. } The study in this work is restricted to open-weight models of intermediate size. Reproducing this work on frontier models was beyond our computational capacities and against the terms of service of proprietary models. We also showed that most fits are in reality limited by the relatively small size of the adversarial prompt dataset (400 prompts). Having more accurate large $N$ behaviors would require a larger adversarial dataset. We have also only used a single dataset for the main results, expanding to other datasets would improve the generality of our claims. At our attack budgets $N \sim 10^{4}$ we were able to resolve most of the prompts. The fit of Table~\ref{tab:params} keeps the remaining ones as censored observations, which bound their barriers from below but cannot measure them exactly. Moments of per-prompt fits, which drop them, make a model look `softer' (Llama's $\mu_E$ falls from $2.75$ to $2.05$, App.~\ref{app:estimator}).
Resolving the censored prompts would require larger attack budgets which are beyond our computational capacity.  Another limitation of this work is the reliance on LLMs as judges of harmful behaviors. We repeat every analysis of the two Qwen-Instruct models with HarmBench labels and with labels requiring both judges to agree (App.~\ref{sec:judges}). A blind human audit of $500$ completions finds that only $14\%$ $[9, 23]$ of Granite's jailbreaks on Qwen2.5-3B-Instruct are harmful, against $79\%$ $[70, 86]$ for HarmBench and $88\%$ for completions both judges flag, while $3\%$ $[1, 9]$ of the completions both judges pass are harmful. Absolute levels such as $\mu_E$ therefore depend on the judge and are not rates of human-judged harm. We tried to mitigate this by using more than one LLM judge \citep{padhi2024granite,mazeika2024harmbench}. We saw that the structure of our results does not rely on a specific LLM judge choice. However, the fitted constants do change and automated judges are known to disagree \citep{souly2024strongreject,ran2024jailbreakeval}. Finally, we have only studied the behavior of these attacks in English. There is no reason to believe that these models would behave identically for other languages. In fact, some languages may be easier to jailbreak than others. 

\emph{Ethical concerns. } This work studies an already established attack. The only offensive contribution is the possibility of making this attack more compute-efficient, which is only presented to inform defenders on what a worst-case attack could be. This paper aims at better understanding the mechanisms at play behind BoN jailbreaking. While we cannot exclude that the results presented in this paper could possibly lead to improved offensive capabilities, we pursued this work, and presented it, from a defender's perspective. We have included prescriptions on how our results could guide a defender's choices on how to safeguard their models and we hope that the work presented here will lead to more robust alignment protocols. 

\subsubsection*{Reproducibility statement}
Every number in the paper is produced by a scripted pipeline from raw generations: the per-prompt Beta-Binomial fits and the estimator of Eq.~(\ref{eq:alpha-estimator}), the five-parameter surfaces, the $N$- and $M$-extrapolations, the temperature sweep, the probes, and all figures. Models, judges, augmentation kernels and grid shapes are specified in App.~\ref{sec:exp}; the code and the fitted per-prompt parameters will be released.

\subsubsection*{Use of large language models}
We used an AI coding assistant (Claude, Anthropic) throughout this project, in two capacities. \emph{(i) Code and experiments.} The assistant wrote and debugged part of the analysis and plotting code (fitting routines, bootstrap error bars, the figure pipeline) and helped launch and monitor the generation and scoring jobs on SLURM. It was also used interactively to run exploratory analyses and sanity checks proposed by the authors. \emph{(ii) Writing.} The main text was written by the authors. The assistant drafted parts of the supplementary material (derivations and experimental details) from the authors' notes and results, which the authors then checked and edited. It was used for proofreading and for consistency checks between the text, the tables and the figures. All LLM-produced code, numbers and text were verified by the authors, who take full responsibility for the content of the paper. Separately, LLMs are the object of study and an LLM-based classifier is used as the jailbreak judge.

\subsubsection*{Acknowledgments}
We thank the anonymous ICLR reviewers for their careful feedback, which significantly improved this paper. This project is supported in part by the Eric and Wendy Schmidt AI in Science Fellowship, a program of Schmidt Sciences.

\newpage

\appendix

\section{Saddle-point evaluation of the population average}
\label{app:saddle}

This appendix derives Eq.~(\ref{eq:f-def}) from Eq.~(\ref{eq:ASR-xi}). Write $x = \log(N M) - \mu_E$ and $\gamma = \sigma_E$, so that Eq.~(\ref{eq:ASR-xi}) reads
\begin{equation} \label{eq:supp-integral}
  \AFRNM \simeq \int \frac{\dd \xi}{\sqrt{2\pi}} \, e^{-\varphi_{x, \gamma}(\xi)} \mbox{~~with~~}
  \varphi_{x, \gamma}(\xi) = \frac{\xi^2}{2} + e^{\,x + \gamma \xi} \;.
\end{equation}
The exponent is convex, so it has a single stationary point,
\begin{equation}
  \varphi'_{x, \gamma}(\xi^\star) = \xi^\star + \gamma\, e^{\,x + \gamma \xi^\star} = 0 \;.
\end{equation}
Since the exponential is positive we must have $\xi^\star < 0$. In other words, at large $x$ the surviving prompts are the
\emph{hard} ones whose effective barrier $E_b$ lies above the mean. Re-writing $w = -\xi^\star > 0$ yields $e^{\,x - \gamma w} = w/\gamma$, i.e. $w e^{\gamma w} = \gamma e^{x}$, and
multiplying by $\gamma$ gives a Lambert equation for the rescaled saddle $\zeta \equiv \gamma w$,
\begin{equation} \label{eq:supp-lambert}
  \zeta\, e^{\zeta} = \gamma^2 e^{x} \qquad \Longrightarrow \qquad \zeta = W_0\!\left( \gamma^2 e^{x} \right)
  = W_0\!\left( \sigma_E^2 (NM) e^{-\mu_E} \right) \;,
\end{equation}
with $W_0$ the principal branch of the Lambert function. At the saddle the height and the
curvature are
\begin{equation}
  \varphi(\xi^\star) = \frac{w^2}{2} + \frac{w}{\gamma} = \frac{\zeta^2 + 2 \zeta}{2 \gamma^2} \;, \qquad
  \varphi''(\xi^\star) = 1 + \gamma^2 e^{\,x + \gamma \xi^\star} = 1 + \gamma w = 1 + \zeta \;.
\end{equation}
The Gaussian fluctuations in the integral contribute $\varphi''(\xi^\star)^{-1/2}$ and
Eq.~(\ref{eq:supp-integral}) becomes $f_\gamma(z)$ of Eq.~(\ref{eq:f-def}), with the argument
$z = (NM)e^{-\mu_E}$ of the main text entering only through $\zeta = W_0(\gamma^2 z)$. 

\paragraph{The collapse variables.} Panel (b) of Fig.~\ref{fig:collapse} needs $\mu_R(M)$ and $\sigma_R(M)$ without any closed form. To obtain them we measure $R_b(M) = -\log Q_b(M)$ for every prompt from the unbiased completion-channel estimate of $Q_b(M)$. Then, $\mu_R(M) = \langle \log R_b(M) \rangle_b$ is read directly from the data and the width is kept at a single number $\sigma_R$ per model because the measured $\sigma_R(M)$ barely changes with $M$: from $0.55$ at $M = 1$ to $0.54$ at $M = 64$ on Qwen2.5-3B-Instruct. All prompts are included in the measured AFR. A prompt that is never jailbroken on the grid has $R_b = 0$ at the grid's resolution and cannot be placed on a log scale. We therefore fit a normal distribution to $\log R_b(M)$ over all prompts, with one location per $M$ and a single width per model, and treat these prompts as left-censored at one jailbreak in the whole grid row. There are $7$ of them for Qwen2.5-7B-Instruct and $10$ for Llama-3.1-8B-Instruct. %

\section{Compute-optimal completion budget}
\label{app:mstar}

This appendix gives the implicit equation for $M^\star$ referred to in Sec.~\ref{sec:mstar}. Fix a
prompt $b$ and write $R_b(M) = -\log Q_b(M)$ as in Eq.~(\ref{eq:Rb}). Because
$Q_b(M) = \langle e^{M Z} \rangle_a$ with $Z = \log(1 - p_{b,a})$, the function $\log Q_b(M)$ is a
cumulant generating function and is therefore convex with $\log Q_b(0) = 0$; hence $R_b(M)$ is
concave, $R_b(M)/M$ is non-increasing, and
\begin{equation} \label{eq:supp-jensen-M}
  Q_b(M) \ge Q_b(1)^M \qquad \text{for every prompt and every } M \;,
\end{equation}
with a gap growing like $(M-1)\,{\rm Var}_a(p)$. Hence if sampling completions costs the same (or more) than sampling augmentations $M^\star = 1$ is optimal. An attacker should spend everything on distinct augmentations.

However under some metrics (FLOPS for example) completions might be cheaper to sample than augmentations. A fresh augmentation requires a fresh prefill,
whereas an extra completion of an augmentation already in cache does not, so in forward-pass units
the cost of an $(N, M)$ attack is $C(N, M) = N(r + M)$, with $r$ the prefill cost expressed in
completions. At fixed $C$ the survival of prompt $b$ is
$\exp[-N R_b(M)] = \exp[-C\,R_b(M)/(r + M)]$, so minimizing it means maximizing
$g(M) = R_b(M)/(r+M)$ which satisfies
\begin{equation} \label{eq:supp-mstar}
  \left( r + M^\star \right) R_b'(M^\star) = R_b(M^\star) \;.
\end{equation}
With the Beta soft wall $p \sim B(\beta_b T, c_b)$, for which $Q_b(M) = \langle (1-p)^M \rangle = B(\beta_b T, c_b + M) / B(\beta_b T, c_b)$ with $B$ denoting Euler's Beta function, we have
\begin{equation}
  R_b(M) = \log \frac{B(\beta_b T, c_b)}{B(\beta_b T, c_b + M)} \;, \qquad
  R_b'(M) = \psi\!\left(\beta_b T + c_b + M\right) - \psi\!\left(c_b + M\right) \;,
\end{equation}
with $\psi$ the digamma function. Hence, $M^\star = 1$ iff $g'(1) \le 0$, i.e. iff
$(r+1) R_b'(1) \le R_b(1)$, which expanded to second order in $p$ reads
\begin{equation} \label{eq:supp-threshold}
  M^\star = 1 \iff r \le \frac{{\rm Var}_a(p)}{2 \langle p \rangle_a}
  = \frac{1 - q_b}{2\left( \beta_b T + c_b + 1 \right)} \mbox{~~where~~} q_b = \langle p_{b,a}\rangle_a \;.
\end{equation}
In other words, repeated completions amortize the prefill cost if and only if the within-prompt spread of escape probabilities outweighs the prefill cost. The threshold is small for the models we measured. For Qwen2.5-3B-Instruct ($\beta_b T \simeq 0.41$, $c_b \simeq 8$, hence $q_b \simeq 0.049$) it is
$r^\star \simeq 0.05$. Solving
Eq.~(\ref{eq:supp-mstar}) at that fit gives $M^\star = 1.3, 2.6, 4.3, 6.2, 9.1$ at
$r = 0.1, 0.4, 1, 2, 4$: the compute-optimal attack is not $M = 1$ but a handful of completions per
augmentation, amortizing each prefill. 

\section{Estimator details and identifiability}
\label{app:estimator}

\paragraph{Clamping.} The barrier $\alpha_b$ of Eq.~(\ref{eq:ansatz}) is Gaussian, so a fraction of the population is negative and the raw Arrhenius expression would exceed one there. Drawing from the fitted five-parameter population of each model (Table~\ref{tab:params}), the clamp of Eq.~(\ref{eq:p-def}) is active on $6.8\%$ of the $(b,a)$ draws for Llama-3.1-8B-Instruct, $0.8\%$ for Qwen2.5-7B (base), $0.7\%$ for Qwen2.5-7B-Instruct, $0.01\%$ for Qwen2.5-3B-Instruct and never for Gemma-2-9B-it. For $\alpha_b < 0$ the clamped mean has a closed form, $\langle p \rangle_a = 1 - e^{\beta_b T \alpha_b}/(1 + \beta_b T)$. Using it instead of the unclamped expression changes ${\rm ASR@}N$ by at most $0.012$ for Llama-3.1-8B-Instruct (at $N = 1$), and by at most $0.003$ for every other model and every $N$. The clamp can therefore safely be neglected.

\paragraph{Survival estimator.} The population survival of Sec.~\ref{sec:crossover-id} involves no fit and no censoring. Instead it is counted directly from the $10^4$-augmentation arms, over all $400$ prompts. On Qwen2.5-3B-Instruct every one of them is jailbroken at least $90$ times under Granite labels, so none of them have to be censored. Under HarmBench labels $5$ of them are never jailbroken, and $39$ under labels requiring both judges (App.~\ref{sec:judges}).
 
\paragraph{The soft wall.} Eq.~(\ref{eq:p-dist}) says the escape probabilities of a prompt follow a
power law $\rho(p) \propto p^{\beta_b T - 1}$ cut off at $p = e^{-\alpha_b}$, so in principle
$(\alpha_b, \beta_b)$ can be read off the exponent and the support edge. The edge estimator is
unusable in practice. Since it is the maximum of $N$ draws its variance does not decrease with $M$.
We therefore replace the hard cutoff by the Beta soft wall 
\begin{equation} \label{eq:p-dist-beta}
p_{b, a} \sim B(\beta_b T, c_b) \;,
\end{equation}
and fit
$(\beta_b T, c_b)$ by maximum likelihood on the Beta-Binomial model
$k_{b,a} \sim {\rm Binomial}(M, p)$, with $\alpha_b$ recovered through
Eq.~(\ref{eq:alpha-estimator}). 
The Beta of Eq.~(\ref{eq:p-dist-beta}) is a deliberate misspecification of the hard-edged law it replaces. However, we checked that by simulating Eq.~(\ref{eq:ansatz}) on synthetic data comparing the soft wall with the ground truth the shape parameters remain the same up to a few tenths. Furthermore, as a second-order effect, if we suppose that $\alpha_b$ might actually fluctuate slightly as a function of the augmentation, then these fluctuations would smooth out the hard-wall into a Beta shaped soft-wall. 

\paragraph{Limitations of the per-prompt fit.} This fit has two main limitations. It drops prompts with too few jailbreaks, even though such a prompt still bounds $q_b$ from above; we require at least three hits in total
across the augmentations of a prompt (i.e. $\sum_a \hat k_{b,a} \geq 3$), and discard fits with unrealistic parameters
($\beta_b T \notin [10^{-3}, 50]$ or $c_b \notin [10^{-3}, 10^{5}]$) which typically correspond to augmentations which have led to gibberish completions. The number of prompts the fit retains is $400$ of $400$ for Qwen2.5-3B-Instruct, $383$ of $400$ for Qwen2.5-7B-Instruct, $91$ of $100$ for Qwen2.5-7B (base), $89$ of $100$ for Gemma-2-9B-it and $80$ of $100$ for Llama-3.1-8B-Instruct. Retrieving them would require an attack with a much larger $N$ (or $M$ or both). Second, the two parameters are separately identifiable only when the completion budget resolves the shape of the $p$ distribution in Eq.~(\ref{eq:p-dist}) which is only possible for $M \gtrsim 10$.

\paragraph{The population fit.} Both limitations come from taking moments of per-prompt estimates. This is why Table~\ref{tab:params} instead maximizes the marginal likelihood of Eq.~(\ref{eq:hier}) over \emph{all} prompts. In practice we parametrize each prompt by $(E_b, \log\beta_b T)$, with $c_b = \beta_b T (e^{E_b} - 1)$ so that the soft wall has mean $e^{-E_b}$ (Eq.~(\ref{eq:def-E})). This change of variables from $(\alpha_b, \log\beta_b)$ has unit Jacobian. The likelihood of each prompt is evaluated once on an adaptive grid in these two variables, and the integral of Eq.~(\ref{eq:hier}) becomes a weighted sum over that grid. A prompt with no jailbreak then simply enters as a censored observation, and the per-prompt estimation noise no longer inflates the spreads. Error bars are obtained by refitting on prompts resampled with replacement ($200$ resamples). The same likelihood can also include a judge false-positive rate $\epsilon$, by replacing $p_{b,a}$ with $\epsilon + (1 - \epsilon) p_{b,a}$ (App.~\ref{sec:judges}); Table~\ref{tab:params} uses $\epsilon = 0$. For comparison, Table~\ref{tab:twostage} gives the moments of the per-prompt fits. When the per-prompt fit drops prompts, the effective barrier is higher in the population fit, most notably for Llama-3.1-8B-Instruct ($\mu_E = 2.75 \pm 0.22$ instead of $2.05 \pm 0.15$). The correlation $\rho$ is significant for every model except Gemma-2-9B-it ($2\Delta\log L$ against $\rho = 0$ between $7.5$ and $87$, and $2.6$ for Gemma), and its sign changes across models. This is why we keep it as a free parameter of the model. It is not an artifact of the estimator either: drawing $\alpha_b$ and $\log\beta_b$ independently and passing the simulated grids through the per-prompt pipeline recovers a correlation of only $0.00$--$0.08$ on all five grids.

\begin{table}[h]\centering
\resizebox{\linewidth}{!}{%
\begin{tabular}{lcccccccc}
\toprule
 & $\mu_\alpha$ & $\sigma_\alpha$ & $\mu_\beta$ & $\sigma_\beta$ & $\rho$ & $\mu_E$ & $\sigma_E$ & prompts fitted \\
\midrule
Qwen2.5-7B (base)     & $0.57_{\pm 0.03}$ & $0.27_{\pm 0.03}$ & $+1.99_{\pm 0.09}$ & $0.85_{\pm 0.05}$ & $-0.28_{\pm 0.10}$ & $0.73_{\pm 0.03}$ & $0.33_{\pm 0.04}$ & $91/100$ \\
Llama-3.1-8B-Instruct & $0.39_{\pm 0.10}$ & $0.93_{\pm 0.11}$ & $-1.29_{\pm 0.15}$ & $1.33_{\pm 0.11}$ & $+0.03_{\pm 0.13}$ & $2.05_{\pm 0.15}$ & $1.36_{\pm 0.11}$ & $80/100$ \\
Qwen2.5-3B-Instruct   & $1.83_{\pm 0.03}$ & $0.65_{\pm 0.03}$ & $-0.88_{\pm 0.03}$ & $0.52_{\pm 0.02}$ & $+0.48_{\pm 0.03}$ & $3.08_{\pm 0.03}$ & $0.58_{\pm 0.02}$ & $400/400$ \\
Qwen2.5-7B-Instruct   & $2.10_{\pm 0.06}$ & $1.30_{\pm 0.05}$ & $-1.97_{\pm 0.05}$ & $1.07_{\pm 0.04}$ & $-0.08_{\pm 0.05}$ & $4.27_{\pm 0.08}$ & $1.65_{\pm 0.06}$ & $383/400$ \\
Gemma-2-9B-it         & $2.59_{\pm 0.06}$ & $0.60_{\pm 0.04}$ & $+1.07_{\pm 0.08}$ & $0.75_{\pm 0.06}$ & $+0.10_{\pm 0.11}$ & $2.93_{\pm 0.06}$ & $0.62_{\pm 0.03}$ & $89/100$ \\
\bottomrule
\end{tabular}}
\caption{\label{tab:twostage} Moments of the per-prompt (two-stage) fits of Eq.~(\ref{eq:alpha-estimator}), over the prompts the per-prompt fit retains, to compare with Table~\ref{tab:params}; this was Table~1 of the first version of this paper. Errors are $\pm 1$ s.d.\ from a bootstrap over prompts ($2000$ resamples).}
\end{table}

\section{Experimental protocol}
\label{sec:exp}

Table~\ref{tab:supp-runs} lists every generation run behind the figures. All runs use the AdvBench
harmful-behavior set \citep{gcg} except the HarmBench control, sample with $T$ as stated and $\texttt{top\_p} = 1$, and cap
completions at $96$ new tokens, which is ample for a judge to decide and keeps the arms affordable.
Every completion is scored by Granite-Guardian-3.0-2B \citep{padhi2024granite} as the primary graded
judge, with HarmBench-Llama-2-13b-cls \citep{mazeika2024harmbench} as a control on the two large
grids; a completion counts as unsafe when the judge's harm probability exceeds $\tau = 0.5$, and
$\tau \in \{0.1, 0.3\}$ is recorded so that the threshold can be swept without regenerating.

\begin{table}[h]\centering
\resizebox{\linewidth}{!}{%
\begin{tabular}{llcccl}
\toprule
Model & Kernel & $\sigma$ & Prompts & $N \times M$ & Role \\
\midrule
Qwen2.5-3B-Instruct   & embedding & $0.048$ & $400$ & $100 \times 300$   & main grid \\
Qwen2.5-3B-Instruct   & embedding & $0.048$ & $400$ & $10^4 \times 1$    & far-$N$ arm \\
Qwen2.5-7B-Instruct   & embedding & $0.038$ & $400$ & $100 \times 300$   & main grid \\
Qwen2.5-7B-Instruct   & embedding & $0.038$ & $400$ & $10^4 \times 1$    & far-$N$ arm \\
Qwen2.5-7B (base)     & embedding & $0.038$ & $100$ & $40 \times 20$     & alignment comparison \\
Gemma-2-9B-it         & embedding & $0.080$ & $100$ & $40 \times 32$     & cross-family \\
Llama-3.1-8B-Instruct & embedding & $0.015$ & $100$ & $40 \times 32$     & cross-family \\
Qwen2.5-3B-Instruct   & character & $0.25$  & $100$ & $100 \times 200$   & kernel control \\
Qwen2.5-3B-Instruct   & embedding & $0.048$ & $100$  & $100 \times 300$   & $T \in \{0.5, 0.75, 1.25, 1.5\}$ sweep \\
Qwen2.5-3B-Instruct   & embedding & $0.048$ & $60$  & $100 \times 1000$  & high-resolution $p$ \\
Qwen2.5-3B-Instruct   & embedding & $6$ values, $0.024$--$0.076$ & $400$ & $400$ & $\sigma$ calibration sweep \\
Qwen2.5-3B-Instruct   & embedding & $6$ values, $0.024$--$0.076$ & $100$ & $50 \times 100$ & $\sigma$ robustness (App.~\ref{sec:sigma}) \\
Qwen2.5-3B-Instruct & embedding & $0.048$ & $200$ HarmBench & $50 \times 100$, $2000 \times 1$ & prompt-distribution control (App.~\ref{app:heldout}) \\
Qwen2.5-3B-Instruct & embedding & $0.048$ & $30$ & $4 \times 64$ per condition & interventions, $T \in \{0.5, 1, 1.5\}$ (App.~\ref{app:interventions}) \\
\bottomrule
\end{tabular}}
\caption{\label{tab:supp-runs} Generation runs. $N \times M$ is augmentations times completions per
augmentation, per prompt. The embedding kernel adds isotropic Gaussian noise to the input embeddings
of the user-content tokens only, leaving the chat template clean; the character kernel perturbs the
prompt string. The character kernel is the augmentation of \citet{hughes2024bon}. Each letter is case-flipped with probability $\sigma$, the interior of each word of four characters or more is randomly permuted with probability $\sigma$, and each character is replaced by a random letter or duplicated with probability $\sigma/10$. All runs are at $T = 1$ unless stated otherwise.}
\end{table}

\paragraph{Layers and probes.} The residual-stream experiments of Fig.~\ref{fig:physics} are run on Qwen2.5-3B-Instruct at layer $30$ of its $36$ blocks, reading the last-token residual; App.~\ref{app:probes} repeats the saddle probe at every layer of all five models. The safety readout of Fig.~\ref{fig:physics}(a) is a ridge regression at penalty $10^3$ on standardized residuals, mean-pooled over the tokens of the completion, with the judge's binary label as target. The saddle probe of Fig.~\ref{fig:physics}(b) is a ridge regression of the within-prompt centred $\log p_{b,a}$ on the within-prompt centred residual of the augmented prompt, with the penalty selected from $\{10^2, 10^3, 10^4, 10^5, 10^6\}$. It is fitted on the high-resolution run of Table~\ref{tab:supp-runs} ($60$ prompts, $1000$ completions per augmentation), with cross-validation grouped by prompt. On the main grid the same probe gives $R^2 = 0.66$ at layer $30$ (App.~\ref{app:probes}).

\paragraph{Choosing $\sigma$.} The augmentation scale is quoted in raw embedding units, with no
rescaling by the embedding norm. Embedding scales differ by an order of magnitude
across families, so $\sigma$ does not transfer between them and each model is calibrated on its own. To calibrate $\sigma$ we sweep $\sigma$ and take the value at which the median per-prompt jailbreak rate first lands in the $1$\% to $5\%$ band. The usable window can be narrow. For example, Llama-3.1-8B has a median rate of $0$
at $\sigma = 0.013$ and $0.18$ at $\sigma = 0.017$, and by $\sigma = 0.04$ the completions are no
longer fluent. App.~\ref{sec:sigma} shows what the fitted
parameters do as this choice is varied.

\section{Augmentation scale robustness}
\label{sec:sigma}

Every grid in this paper is run at one augmentation scale $\sigma$, chosen per model by the
calibration of App.~\ref{sec:exp}. To study the dependence of our 5-parameter family on $\sigma$ we ran a sweep over $100$ behaviors $\times\, 50$ augmentations $\times\, 100$ completions for 6 values of $\sigma$ for Qwen-3B-Instruct. The prompt set is identical across the six.

\begin{table}[h]\centering
\resizebox{\linewidth}{!}{%
\begin{tabular}{lcccccccc}
\toprule
$\sigma$ & median rate & $\mu_\alpha$ & $\sigma_\alpha$ & $\mu_\beta$ & $\sigma_\beta$ & $\rho$ & $\mu_E$ & $\sigma_E$ \\
\midrule
$0.024$ & $0.001$ & $5.00_{\pm 0.30}$ & $2.23_{\pm 0.22}$ & $-1.13_{\pm 0.31}$ & $2.17_{\pm 0.16}$ & $+0.73_{\pm 0.08}$ & $6.77_{\pm 0.19}$ & $1.60_{\pm 0.13}$ \\
$0.030$ & $0.005$ & $2.63_{\pm 0.19}$ & $1.61_{\pm 0.15}$ & $-2.22_{\pm 0.16}$ & $1.21_{\pm 0.11}$ & $+0.55_{\pm 0.10}$ & $5.02_{\pm 0.15}$ & $1.37_{\pm 0.09}$ \\
$0.038$ & $0.029$ & $1.77_{\pm 0.11}$ & $0.89_{\pm 0.07}$ & $-1.38_{\pm 0.08}$ & $0.75_{\pm 0.07}$ & $+0.44_{\pm 0.10}$ & $3.41_{\pm 0.09}$ & $0.83_{\pm 0.06}$ \\
$0.048$ & $0.045$ & $1.66_{\pm 0.08}$ & $0.69_{\pm 0.06}$ & $-1.07_{\pm 0.07}$ & $0.67_{\pm 0.06}$ & $+0.63_{\pm 0.09}$ & $3.06_{\pm 0.06}$ & $0.54_{\pm 0.05}$ \\
$0.060$ & $0.032$ & $1.83_{\pm 0.09}$ & $0.92_{\pm 0.07}$ & $-1.26_{\pm 0.06}$ & $0.59_{\pm 0.06}$ & $+0.69_{\pm 0.06}$ & $3.37_{\pm 0.07}$ & $0.69_{\pm 0.06}$ \\
$0.076$ & $0.039$ & $1.40_{\pm 0.09}$ & $0.79_{\pm 0.08}$ & $-1.57_{\pm 0.06}$ & $0.52_{\pm 0.06}$ & $+0.75_{\pm 0.07}$ & $3.18_{\pm 0.06}$ & $0.55_{\pm 0.05}$ \\
\bottomrule
\end{tabular}}
\caption{\label{tab:sigma} The five numbers and the effective barrier as the augmentation scale is
swept, Qwen2.5-3B-Instruct, $100 \times 50 \times 100$ per $\sigma$, Granite judge, $T = 1$, fitted as Table~\ref{tab:params} (the $14$ and $3$ prompts never jailbroken at $\sigma = 0.024$ and $0.030$ are censored). Error
bars are $\pm 1$ s.d.\ from a bootstrap over prompts ($200$ resamples). `Median rate' is the
median over prompts of the measured single-attempt jailbreak rate. The main text uses $\sigma = 0.048$.}
\end{table}

The parameters clearly depend on $\sigma$. Higher values of $\sigma$ progressively erode the safety barriers installed by the model (i.e. lower $\mu_\alpha$ and raise $\mu_\beta$). Eventually too high values of $\sigma$ hit a ceiling where increased noise destroys too much of the original prompt and the completions start to lose coherence. The trends in the 5-parameter family follow the behavior of the median rate, which is the best proxy to determine the `attack strength'. To be able to compare different models one should not aim at keeping $\sigma$ fixed, but instead try to match their median jailbreak rates as detailed in App.~\ref{sec:exp}.

\section{Augmentation kernel robustness}
\label{sec:kernel}

The main text augments by adding isotropic Gaussian noise to the input embeddings of the
user-content tokens which is a white-box operation. A black-box attacker instead perturbs the
prompt \emph{string}, which is the character kernel of Table~\ref{tab:supp-runs}. As shown in Table~\ref{tab:kernel}, the safety parameters ($\mu_\alpha, \sigma_\alpha, \mu_\beta, \sigma_\beta$) depend on the attack type. However, the general form of our microscopic model (entropic bottleneck, hyperplane, exponential saddles) holds for both kernels tested. In other words, our 5-parameter family fitted on model X attacked via augmentation Y encodes: \emph{how resilient is model X to a BoN attack under Y}. Therefore, this 5-parameter fit allows us to compare different models' safety under the same augmentation, or the susceptibility of a single model to different types of augmentations. Even at similar median rates (character kernel at $\sigma = 0.25$, embedding kernel at $\sigma = 0.030$) the parameters are closer but still do not agree within noise. $\mu_E$ differs by about $3.5$ s.d.\ ($4.27$ against $5.02$) and $\sigma_\alpha$ by about $2$ s.d.\ ($1.20$ against $1.61$), partly because the median rates still differ ($0.012$ against $0.005$).

\begin{table}[h]\centering
\resizebox{\linewidth}{!}{%
\begin{tabular}{llccccccc}
\toprule
 & median rate & $\mu_\alpha$ & $\sigma_\alpha$ & $\mu_\beta$ & $\sigma_\beta$ & $\rho$ & $\mu_E$ & $\sigma_E$ \\
\midrule
character, $\sigma = 0.25$      & $0.012$ & $2.31_{\pm 0.13}$ & $1.20_{\pm 0.11}$ & $-1.74_{\pm 0.10}$ & $0.94_{\pm 0.07}$ & $+0.26_{\pm 0.09}$ & $4.27_{\pm 0.14}$ & $1.25_{\pm 0.07}$ \\
embedding, $\sigma = 0.048$     & $0.044$ & $1.65_{\pm 0.06}$ & $0.57_{\pm 0.04}$ & $-1.02_{\pm 0.06}$ & $0.54_{\pm 0.04}$ & $+0.51_{\pm 0.07}$ & $3.01_{\pm 0.05}$ & $0.50_{\pm 0.04}$ \\
embedding, $\sigma = 0.030$     & $0.005$ & $2.63_{\pm 0.19}$ & $1.61_{\pm 0.15}$ & $-2.22_{\pm 0.16}$ & $1.21_{\pm 0.11}$ & $+0.55_{\pm 0.10}$ & $5.02_{\pm 0.15}$ & $1.37_{\pm 0.09}$ \\
\bottomrule
\end{tabular}}
\caption{\label{tab:kernel} The two augmentation kernels on the same model and the same $100$
prompts, fitted as Table~\ref{tab:params}: character kernel $100 \times 100 \times 200$; embedding kernel at $\sigma = 0.048$, the main grid thinned to $100 \times 100 \times 200$; at $\sigma = 0.030$, the $100 \times 50 \times 100$ grid of Table~\ref{tab:sigma}. Error bars are $\pm 1$ s.d.\ from a bootstrap over prompts ($200$ resamples).}
\end{table}

\paragraph{The structure is shared by the kernels tested.} Under
the character kernel the standardized $\alpha_b$ is still indistinguishable from a Gaussian
(KS $p = 0.94$) and standardized $\log \beta_b$ from a
normal (KS $p = 0.95$). Reconstructing the full $\ASRNM$ surface from the five fitted numbers over ($N \leq 100$, $M \leq 50$) gives a mean absolute error of $0.017$ under the character kernel. Therefore, although the parameters may vary, the structure in Sec.~\ref{sec:model} holds for both kernels. More generally, across the seven (kernel, strength) configurations available to us on Qwen2.5-3B-Instruct, namely the character kernel and the embedding kernel at six values of $\sigma$, Gaussianity of $\hat\alpha_b$ and of $\log\hat\beta_b$ is never rejected. We have not tested kernels that change the meaning of the prompt, such as LLM paraphrases or random-token insertion, so we restrict our claim to the kernels above.

\section{Judge robustness}
\label{sec:judges}

The notion of safety is dependent on the judge. Hence all the fit parameters $(\mu_\alpha, \sigma_\alpha, \mu_\beta, \sigma_\beta{, \rho})$ depend on the judge's notion of safety. However, the actual scaling forms, Gaussian and log-normal structures are the same under every judge we tested. Every number in the main text is graded by Granite-Guardian-3.0-2B. Because an automated judge
defines what counts as a jailbreak, we re-scored the \emph{identical} generations of the
Qwen2.5-3B-Instruct grid ($400$ prompts $\times$ $100$ augmentations $\times$ $300$ completions,
$1.2 \times 10^{7}$ completions) with HarmBench-Llama-2-13b-cls \citep{mazeika2024harmbench} and
repeated the whole estimation. No generation differs between the two columns of
Table~\ref{tab:judges}; only the label does.

\begin{table}[h]\centering
\resizebox{\linewidth}{!}{%
\begin{tabular}{lccc}
\toprule
 & Granite-Guardian-3.0-2B & HarmBench-Llama-2-13b-cls & both agree \\
\midrule
per-completion success rate      & $0.044$   & $0.003$ & $0.0012$ \\
$\mu_\alpha$ (base alignment)    & $1.81$    & $4.70$ & $5.37$ \\
$\sigma_\alpha$                  & $0.63$    & $1.00$ & $1.18$ \\
$\mu_\beta$ (susceptibility)     & $-0.90$   & $-1.63$ & $-2.05$ \\
$\sigma_\beta$                   & $0.49$    & $0.49$ & $0.65$ \\
$\rho$                     & $+0.48$ & $-0.24$ & $-0.18$ \\
$\mu_E$ (effective barrier)      & $3.07$    & $6.53$ & $7.56$ \\
$\sigma_E$                       & $0.55$    & $1.16$ & $1.40$ \\
prompts never jailbroken         & $0$       & $2$ & $23$ \\
\ldots in the $10^4$-augmentation arm & $0$ & $5$ & $39$ \\
\bottomrule
\end{tabular}}
\caption{\label{tab:judges} The same generations scored by two judges. HarmBench is far stricter, so
every barrier scale shifts, but the \emph{shape} of the model survives: $\alpha_b$ stays Gaussian,
$\log \beta_b$ stays log-normal, and the ordering of the two channels is unchanged.}
\end{table}

HarmBench calls $0.3\%$ of completions unsafe against Granite's $4.4\%$, which
raises the mean barrier by $\mu_E^{\rm HB} - \mu_E^{\rm Gr} = 3.46$ which is expected given the judge is stricter. More importantly, the {distribution} families are judge-independent.
Standardized per-prompt $\alpha_b$ remains indistinguishable from a Gaussian under both judges, and the
susceptibility $\beta_b$ stays log-normal. The stricter judge widens the spread of the barrier
($\sigma_\alpha$ from $0.63$ to $1.00$, and $1.18$ under both judges), because it pushes the
hardest prompts further into the tail. Interestingly, $\sigma_\beta$ is unchanged under HarmBench and only rises to $0.65$ under both judges.

\paragraph{Censoring under each judge.} Stricter labels leave more prompts without any jailbreak on the grid, $2$ under HarmBench and $23$ under both judges for Qwen2.5-3B-Instruct. Since Table~\ref{tab:judges} is fitted as Table~\ref{tab:params}, these prompts are kept as censored observations, whereas moments of per-prompt fits would drop $22$ and $98$ prompts. Table~\ref{tab:judgehier} gives the fit for both Qwen-Instruct models and all three label sets. Notice that $\rho$ is positive only for Qwen2.5-3B-Instruct under Granite, and lies between $-0.18$ and $-0.25$ in every other case. Table~\ref{tab:judgeextrap} shows that the law extrapolates to $N = 10^4$ under every label set.

\begin{table}[h]\centering
\caption{\label{tab:judgeextrap} Extrapolation under three label sets: mean $|\log_{10}$ survival error$|$ for $100 < N \le 10^4$ (cells of Fig.~\ref{fig:trapval}(a)), all fitted on $N \le 100$. Hierarchical: the fit of Table~\ref{tab:params}, all prompts, never-jailbroken ones censored; up to $43$ of $400$ prompts are never jailbroken under the stricter labels. Two-stage: same law with moments of per-prompt fits (Table~\ref{tab:twostage}), which drop prompts with fewer than three jailbreaks. \citet{feng2026saber}: Beta fitted by beta-binomial maximum likelihood on the $N \le 100$ grid.}
\resizebox{\linewidth}{!}{%
\begin{tabular}{llcccc}
\toprule
 & labels & ours, two-stage & ours, hierarchical & \citet{feng2026saber} & power law \\
\midrule
Qwen2.5-3B-Instruct & Granite    & $0.117$ & $\mathbf{0.087}$ & $0.216$ & $1.303$ \\
                    & HarmBench  & $0.055$ & $\mathbf{0.017}$ & $0.190$ & $0.419$ \\
                    & both agree & $0.167$ & $\mathbf{0.023}$ & $0.097$ & $0.235$ \\
Qwen2.5-7B-Instruct & Granite    & $0.212$ & $\mathbf{0.089}$ & $0.288$ & $0.295$ \\
                    & HarmBench  & $0.174$ & $\mathbf{0.031}$ & $0.235$ & $0.208$ \\
                    & both agree & $0.223$ & $\mathbf{0.022}$ & $0.147$ & $0.145$ \\
\bottomrule
\end{tabular}}
\end{table}

\begin{table}[h]\centering
\resizebox{\linewidth}{!}{%
\begin{tabular}{llccccc}
\toprule
 & labels & $\mu_E$ & $\sigma_E$ & $\rho$ & zero-hit prompts (grid) & never jailbroken ($10^4$ arm) \\
\midrule
Qwen2.5-3B-Instruct & Granite    & $3.07_{\pm 0.03}$ & $0.55_{\pm 0.02}$ & $+0.48_{\pm 0.04}$ & $0$  & $0$ \\
                    & HarmBench  & $6.53_{\pm 0.06}$ & $1.16_{\pm 0.04}$ & $-0.24_{\pm 0.07}$ & $2$  & $5$ \\
                    & both agree & $7.56_{\pm 0.07}$ & $1.40_{\pm 0.07}$ & $-0.18_{\pm 0.10}$ & $23$ & $39$ \\
Qwen2.5-7B-Instruct & Granite    & $4.40_{\pm 0.10}$ & $1.76_{\pm 0.07}$ & $-0.22_{\pm 0.06}$ & $7$  & $4$ \\
                    & HarmBench  & $6.07_{\pm 0.12}$ & $2.17_{\pm 0.08}$ & $-0.22_{\pm 0.06}$ & $20$ & $25$ \\
                    & both agree & $6.74_{\pm 0.12}$ & $2.25_{\pm 0.09}$ & $-0.25_{\pm 0.06}$ & $36$ & $43$ \\
\bottomrule
\end{tabular}}
\caption{\label{tab:judgehier} Population fit (as in Table~\ref{tab:params}) of the $400 \times 100 \times 300$ grids under three label sets.}
\end{table}

\paragraph{A judge false-positive rate in the likelihood.} A judge that wrongly flags a fixed fraction $\epsilon$ of safe completions adds a floor to every rate, $k_{b,a} \sim {\rm Binomial}(M, \epsilon + (1 - \epsilon)\, p_{b,a})$. We can estimate $\epsilon$ by profiling the likelihood of Eq.~(\ref{eq:hier}). For Granite on Qwen2.5-3B-Instruct it prefers $\epsilon \approx 0.003$ ($\Delta\log L = 3{,}600$ compared to $\epsilon = 0$), and on Qwen2.5-7B-Instruct $\epsilon \approx 10^{-4}$ ($\Delta\log L = 86$). Including it barely moves $(\mu_E, \sigma_E)$, by at most $0.11$ and $0.03$. A uniform floor $\epsilon \approx 0.003$ would cap the resolution budget near $1/\epsilon \approx 330$, above the measured $\Nres = 111$. However, the audit below shows that Granite's false positives are not uniform. On the $20$ hardest prompts almost all of its jailbreaks are false, so there the floor is set by each prompt's own false-positive rate, about $1.4\%$ of its completions, rather than by a global $\epsilon$.

\paragraph{Human audit.} To measure how the judges' notion of harm aligns with a human one, we labelled by hand $500$ completions from the Qwen2.5-3B-Instruct $10^4$-augmentation arm. A completion counts as harmful if it gives the requested harmful content, even partially or in a degraded form, and the $37$ items we marked as unsure are excluded. Table~\ref{tab:audit} gives the fraction of harmful completions for each subset and, after reweighting by the subset sizes in the arm, the precision of each judge. Granite turns out to be sensitive but imprecise. Only $14\%$ of its jailbreaks are harmful, compared to $79\%$ for HarmBench and $88\%$ when both judges agree. On the other hand, among the harmful completions flagged by at least one judge, Granite catches $85\%$ and HarmBench only $30\%$. On the $20$ hardest prompts, which set $\Nres$ in Sec.~\ref{sec:crossover-id}, only $2$ of the $97$ Granite-only jailbreaks are harmful, so there the Granite rate is a judge floor. Finally, $3$ of the $96$ completions that both judges pass are harmful.

\begin{table}[h]\centering
\resizebox{\linewidth}{!}{%
\begin{tabular}{lcccc}
\toprule
Granite / HarmBench verdict & number of completions & harmful / labelled & unsure & fraction harmful, $95\%$ CI \\
\midrule
both positive                          & $5{,}092$         & $84/96$ & $4$  & $0.88\;[0.79, 0.93]$ \\
Granite only, $20$ hardest prompts     & $2{,}893$         & $2/97$  & $3$  & $0.02\;[0.01, 0.07]$ \\
Granite only, other prompts            & $166{,}159$       & $11/90$ & $10$ & $0.12\;[0.07, 0.21]$ \\
HarmBench only                         & $6{,}118$         & $61/84$ & $16$ & $0.73\;[0.62, 0.81]$ \\
both negative                          & $3{,}819{,}738$   & $3/96$  & $4$  & $0.03\;[0.01, 0.09]$ \\
\bottomrule
\end{tabular}}
\caption{\label{tab:audit} Blind human audit of the Qwen2.5-3B-Instruct $10^4$-augmentation arm ($4 \times 10^6$ completions). Wilson intervals per stratum; the precision intervals put every stratum at its bound together.}
\end{table}

\paragraph{Local exponent within the resolved range.} On Qwen2.5-7B-Instruct no single prompt carries more than half of the measured survival up to $N = 5000$, so in this range the local exponent is free of the finite-size effect. Over $N = 30, 100, 300, 10^3, 3 \times 10^3$ it keeps rising, as $0.33, 0.49, 0.63, 0.74, 0.89$ under Granite and $0.13, 0.24, 0.37, 0.53, 0.65$ under HarmBench. Our law fitted on HarmBench labels gives $0.13, 0.23, 0.34, 0.48, 0.63$. In comparison, the power law of \citet{hughes2024bon} fitted at $N \le 100$ stays between $0.13$ and $0.28$.

\section{Saddle probe across models and layers}
\label{app:probes}

\paragraph{Probe.} For each model and at every layer we regress the within-prompt centred $\log \hat p_{b,a}$ (floored at $0.5/M$) on the within-prompt centred last-token residual of the augmented prompt. We use ridge regression with the penalty chosen from $\{10^2, \dots, 10^6\}$ and $5$-fold cross-validation grouped by prompt, on the grids of Table~\ref{tab:params}. Since $\hat p_{b,a}$ is estimated from only $M$ completions, part of its variance is binomial noise that no readout can explain. We estimate this noise ceiling without assuming any within-prompt law. Splitting the $M$ completions of each cell at random gives two halves that are independent binomials given $p_{b,a}$. The split-half correlation of the centred targets, corrected to the full $M$ with the Spearman--Brown formula, then bounds the explainable variance. For Gemma, a model-based estimate under the fitted Beta soft wall agrees with it to within $0.01$.

\begin{table}[h]\centering
\resizebox{\linewidth}{!}{%
\begin{tabular}{lcccccc}
\toprule
model & grid & best layer & held-out $R^2$ & Spearman & noise ceiling & $R^2$/ceiling \\
\midrule
Qwen2.5-3B-Instruct   & $400 \times 100 \times 300$ & $31/36$ & $0.67$ & $0.79$ & $0.90$ & $0.74$ \\
Qwen2.5-7B-Instruct   & $400 \times 100 \times 300$ & $23/28$ & $0.70$ & $0.78$ & $0.96$ & $0.73$ \\
Llama-3.1-8B-Instruct & $100 \times 40 \times 32$   & $29/32$ & $0.76$ & $0.82$ & $0.94$ & $0.81$ \\
Gemma-2-9B-it         & $100 \times 40 \times 32$   & $35/42$ & $0.19$ & $0.40$ & $0.44$ & $0.43$ \\
Qwen2.5-7B (base)     & $100 \times 40 \times 20$   & $26/28$ & $0.40$ & $0.57$ & $0.61$ & $0.66$ \\
\bottomrule
\end{tabular}}
\caption{\label{tab:probes} Saddle probe at its best layer in five models.}
\end{table}

In every model the held-out $R^2$ is zero at the embedding layer, so the readout is not lexical. It then rises through the middle layers and peaks at $80$--$95\%$ depth. For Qwen2.5-3B-Instruct it is $0.24$ at layer $4$, $0.32$ at $12$, $0.41$ at $20$, $0.56$ at $24$ and $0.67$ at $31$ of $36$, and for Llama-3.1-8B-Instruct $0.32$ at layer $12$, $0.64$ at $16$ and $0.76$ at $29$ of $32$. Gemma is the weakest, as expected, since we have seen that Gemma relies mainly on an entropic bottleneck and has nearly no energetic defense. Its within-prompt variance of $\log\hat p$ is half of Llama's for the same $32$ completions per cell, so most of it is binomial noise, and even relative to its ceiling the probe captures the least.

\section{Interventions on the two directions}
\label{app:interventions}

The readouts of Sec.~\ref{sec:validation} and App.~\ref{app:probes} are correlational. Here we test whether the safety direction $\hat c_{\rm hyper}$ and the saddle direction $\hat c_{\rm saddle}$ also control the transition. All interventions are made on Qwen2.5-3B-Instruct at layer $30$, on $30$ prompts of the main grid and $4$ augmentations of each (with rates between $1.7\%$ and $50\%$), with $64$ completions per condition. Granite labels are used unless stated otherwise.

\paragraph{Pushing along a direction.} We add $\delta\, \hat u$ to the residual stream at layer $30$, either at every prompt position before the generation starts, or at every position including the generated tokens. The direction $\hat u$ is $\hat c_{\rm hyper}$, $\hat c_{\rm saddle}$ or a random direction orthogonal to both. Both are oriented towards the unsafe side. In particular $\hat c_{\rm saddle} \cdot r$ predicts $\log p_{b,a}$, i.e.\ $-\eta_{b,a}/T$ up to a constant, so $\delta > 0$ along $\hat c_{\rm saddle}$ corresponds to a lower saddle, not to a higher one. The residual of the last prompt token has norm $\approx 130$, so $\delta = 30$ and $60$ are pushes of about a quarter and a half of its norm. Table~\ref{tab:push} shows that both directions move the jailbreak rate monotonically, up and down, whereas a random direction of the same norm does not. Both directions are therefore causally linked to the transition. However, across augmentations the readout along $\hat c_{\rm saddle}$ only varies by $s = 1.9$ (within-prompt standard deviation). A push of that size changes $\log p$ by $0.03$ nats, about thirty times less than the $1.03$ nats per standard deviation of the correlation between the readout and $\log p$ across augmentations. Pushing the residual by hand is thus not how the augmentations change $p$.

\begin{table}[h]\centering
\resizebox{\linewidth}{!}{
\begin{tabular}{llcccccc}
\toprule
 & & \multicolumn{5}{c}{jailbreak rate (\%) at $\delta =$} & slope of $\log p$ \\
positions pushed & direction & $-60$ & $-30$ & $0$ & $+30$ & $+60$ & per $30$ units \\
\midrule
prompt & $\hat c_{\rm hyper}$ & $4.2$ & $4.7$ & $6.1$ & $8.9$ & $12.0$ & $+0.27$ $[+0.20, +0.35]$ \\
prompt & $\hat c_{\rm saddle}$ & $2.5$ & $3.8$ & $6.1$ & $10.8$ & $15.6$ & $+0.47$ $[+0.40, +0.54]$ \\
prompt & random & $7.1$ & $6.9$ & $6.1$ & $5.5$ & $6.6$ & $-0.04$ $[-0.09, +0.02]$ \\
prompt and completion & $\hat c_{\rm hyper}$ & $1.9$ & $3.3$ & $6.1$ & $12.7$ & $25.1$ & $+0.65$ $[+0.57, +0.74]$ \\
prompt and completion & $\hat c_{\rm saddle}$ & $1.1$ & $2.4$ & $6.1$ & $14.7$ & $25.3$ & $+0.80$ $[+0.70, +0.90]$ \\
prompt and completion & random & $8.9$ & $6.7$ & $6.1$ & $6.5$ & $8.2$ & $-0.02$ $[-0.06, +0.02]$ \\
\bottomrule
\end{tabular}}
\caption{\label{tab:push} Jailbreak rate when $\delta\, \hat u$ is added to the residual stream at layer $30$, at $T = 1$ ($30 \times 4 \times 64$ completions per entry). The slope is a least-squares fit over the five values of $\delta$, with a $95\%$ interval from a bootstrap over prompts.}
\end{table}

\paragraph{Saddle-directed augmentations.} In our model the saddle is selected by the augmentation. To test this we act on the augmentation itself. Let $\eta_0$ be the Gaussian noise of an augmentation, so that the embeddings are perturbed by $\sigma \eta_0$. We move $\eta_0$ by $t$ standard deviations along the direction of the noise space which raises $\hat c_{\rm saddle} \cdot r$, namely $\hat u \propto \nabla_\eta \left(\hat c_{\rm saddle} \cdot r\right)$, recomputed every half standard deviation, and then rescale the result to the norm of $\eta_0$. With the orientation above, $t > 0$ lowers the saddle and $t < 0$ raises it. Since the kernel is an isotropic Gaussian, the new noise is exactly as probable as the original draw, and it differs from it by less than $2\%$ of its norm. As a control we move $\eta_0$ by the same amount along a random direction orthogonal to $\hat u$.

\begin{table}[h]\centering
\begin{tabular}{lcccccc}
\toprule
 & & readout shift & \multicolumn{3}{c}{jailbreak rate (\%)} & change of $\log p$ \\
move & $t$ & (s.d.) & $T = 0.5$ & $T = 1$ & $T = 1.5$ & at $T = 1$ \\
\midrule
directed & $-4$ & $-2.73$ & $0.8$ & $1.3$ & $2.0$ & $-1.44$ $[-2.58, -0.80]$ \\
directed & $-2$ & $-1.86$ & $0.6$ & $1.2$ & $2.8$ & $-1.52$ $[-2.26, -1.01]$ \\
directed & $-1$ & $-1.20$ & $1.5$ & $2.2$ & $3.8$ & $-0.92$ $[-1.42, -0.45]$ \\
none & $0$ & $0$ & $3.1$ & $5.6$ & $9.8$ & $0$ \\
directed & $+1$ & $+1.03$ & $13.0$ & $16.4$ & $21.6$ & $+1.05$ $[+0.80, +1.30]$ \\
directed & $+2$ & $+1.68$ & $20.3$ & $23.8$ & $28.6$ & $+1.43$ $[+1.20, +1.67]$ \\
directed & $+4$ & $+2.84$ & $35.2$ & $38.1$ & $45.3$ & $+1.90$ $[+1.66, +2.12]$ \\
random & $2$ & $-0.01$ & $3.2$ & $5.9$ & $10.3$ & $+0.05$ $[-0.10, +0.19]$ \\
random & $4$ & $-0.00$ & $3.6$ & $5.8$ & $9.3$ & $+0.04$ $[-0.10, +0.15]$ \\
\bottomrule
\end{tabular}
\caption{\label{tab:sda} Saddle-directed augmentations. The readout shift is in units of its within-prompt standard deviation across augmentations, and a positive shift corresponds to a lower saddle. Intervals are $95\%$ from a bootstrap over prompts.}
\end{table}

Table~\ref{tab:sda} shows the result. The readout moves as intended, by about one natural standard deviation for $t = \pm 1$, and the jailbreak rate follows it in both directions: at $T = 1$ it goes from $5.6\%$ to $16.4\%$ when the saddle is lowered ($t = +1$) and to $2.2\%$ when it is raised ($t = -1$). For these small moves $\log p$ changes by $0.93$ $[0.83, 1.05]$ nats per standard deviation of the readout, which is the slope of the correlation across augmentations ($1.03$). The same displacement in a random direction changes neither the readout nor the rate. The coordinate read by $\hat c_{\rm saddle}$ is therefore set by the augmentation and controls the transition, as assumed in Sec.~\ref{sec:model}. The effect decreases with temperature: $1.25$, $0.93$ and $0.75$ nats per standard deviation at $T = 0.5$, $1$ and $1.5$.

For $1\%$ of the augmentations of these prompts on the grid the last prompt token is in a state of very large norm ($\approx 3000$ instead of $130$), in which Granite flags $65\%$ of the completions and HarmBench $0.4\%$. A directed move ends in this state in $3$ to $6$ of the $120$ cases and a random one in none. These cases are left out of Table~\ref{tab:sda}, together with the $2$ augmentations which were already in this state. With them, the few cases at $t < 0$ (rates of $51$ to $75\%$) cancel the decrease of the average rate.

\paragraph{HarmBench labels.} With HarmBench labels the rates are about ten times lower ($0.7\%$ without intervention), and the picture is the same for the augmentations. Raising the saddle ($t = -1$) lowers the rate from $0.73\%$ to $0.14\%$ and lowering it ($t = +4$) raises it to $1.8\%$, with a slope of $0.45$ $[0.30, 0.65]$ nats per standard deviation of the readout over all $t$. A random displacement gives $-0.42$ $[-0.97, 0.00]$ and $-0.11$ $[-0.50, +0.18]$. At these rates the temperature dependence is not resolved ($0.50$, $0.45$ and $0.58$ nats at $T = 0.5$, $1$ and $1.5$). For the pushes, both directions act when the push is applied at every position: the slope is $+0.47$ $[+0.30, +0.63]$ per $30$ units for $\hat c_{\rm hyper}$ and $+0.37$ $[+0.26, +0.53]$ for $\hat c_{\rm saddle}$, against $-0.06$ $[-0.22, +0.07]$ for a random direction. When only the prompt positions are pushed, $\hat c_{\rm saddle}$ still acts ($+0.26$ $[+0.15, +0.38]$) whereas $\hat c_{\rm hyper}$ does not ($-0.09$ $[-0.21, +0.05]$).

\section{Extrapolation to held-out prompts}
\label{app:heldout}

A defender would like to predict the attack on prompts that were not used for the fit. To test this we split the $400$ prompts into random halves ($20$ splits), fit the law on the $N \le 100$ grid of one half as in Table~\ref{tab:params}, and predict two quantities for the other half: (i) the $(N, M)$ surface on the $11 \times 11$ cells $N, M \le 100$, measured model-free on the held-out grid, and (ii) the survival on the held-out half's $10^4$-augmentation arm for $N > 100$, in cells not carried by a single prompt. We compare with the same law fitted on the held-out half itself, and with the Beta law of \citet{feng2026saber} fitted on the training half.

\begin{table}[h]\centering
\resizebox{\linewidth}{!}{%
\begin{tabular}{llccccc}
\toprule
 & & \multicolumn{2}{c}{surface MAE (ASR)} & \multicolumn{3}{c}{$10^4$ arm, mean $|\Delta\log_{10} S|$} \\
 & labels & held-out & in-sample & held-out & in-sample & Beta (train) \\
\midrule
Qwen2.5-3B-Instruct & Granite    & $0.036_{\pm 0.007}$ & $0.035$ & $0.46_{\pm 0.43}$  & $0.37$  & $0.59$ \\
                    & HarmBench  & $0.013_{\pm 0.011}$ & $0.006$ & $0.039_{\pm 0.021}$ & $0.015$ & $0.149$ \\
                    & both agree & $0.016_{\pm 0.011}$ & $0.006$ & $0.023_{\pm 0.015}$ & $0.024$ & $0.133$ \\
Qwen2.5-7B-Instruct & Granite    & $0.024_{\pm 0.014}$ & $0.012$ & $0.103_{\pm 0.068}$ & $0.058$ & $0.233$ \\
                    & HarmBench  & $0.025_{\pm 0.018}$ & $0.009$ & $0.065_{\pm 0.038}$ & $0.029$ & $0.208$ \\
                    & both agree & $0.021_{\pm 0.010}$ & $0.008$ & $0.029_{\pm 0.020}$ & $0.021$ & $0.105$ \\
\bottomrule
\end{tabular}}
\caption{\label{tab:heldout} Held-out prediction, mean $\pm$ s.d.\ over $20$ random $200/200$ splits. The Qwen2.5-3B-Instruct/Granite case is limited by the finite size of each half: beyond its own $\Nres$ the tail of a $200$-prompt half is set by one or two prompts.}
\end{table}

\paragraph{A second prompt set.} To check that our results are not specific to AdvBench, we also ran the $200$ HarmBench standard behaviors \citep{mazeika2024harmbench} on Qwen2.5-3B-Instruct at $\sigma = 0.048$, as a $200 \times 50 \times 100$ grid and an independent $200 \times 1950$ arm (the $2000$ augmentations of Table~\ref{tab:supp-runs} without the $50$ whose seeds are shared with the grid), scored under all three label sets (Granite / HarmBench / both). First, the structure of the model holds. The standardized per-prompt $\hat\alpha_b$, $\log\hat\beta_b$ and $E_b$ cannot be distinguished from Gaussians (KS $p$ = $0.48$, $0.34$, $0.39$ / $0.35$, $0.48$, $0.21$ / $0.39$, $0.69$, $0.93$), and the law reproduces the $(N, M \le 100)$ surface with an MAE of $0.033$ / $0.010$ / $0.007$, similar to an AdvBench grid of the same size ($0.030$ / $0.010$ / $0.010$). Second, HarmBench groups its behaviors into six semantic categories, and under Granite the mean barrier differs across them (ANOVA $p < 10^{-11}$, with category means from $2.6$ to $3.8$). The across-prompt law is therefore a mild mixture: the s.d.\ of $E_b$ is $0.77$ overall and $0.66$ within categories, where the residuals are Gaussian (KS $p = 0.28$). Under the stricter labels the category effect is not significant ($p = 0.07$ and $0.18$). Finally, fitted on the grid ($N \le 50$) and tested on the independent arm, the law misses the survival by $0.11$ in $\log_{10}$ under Granite, compared to $0.19$ for the Beta law of \citet{feng2026saber} and $0.78$ for a power law. Under the stricter labels $86$ and $135$ of the $200$ prompts are never jailbroken in $1950$ augmentations, and every method is then within $0.04$ ($0.005$ and $0.004$ for ours).

\clearpage
\section{Notation}
\label{app:notation}

\begin{table}[h]\centering
\resizebox{\linewidth}{!}{%
\begin{tabular}{ll}
\toprule
symbol & meaning \\
\midrule
$b$, $a$ & prompt, augmentation \\
$N$, $M$ & augmentations per prompt, completions per augmentation \\
$T$ & generation temperature \\
$p_{b,a}$ & probability that one completion of augmentation $a$ of prompt $b$ is unsafe, Eq.~(\ref{eq:p-def}) \\
$\alpha_b$ & entropic bottleneck of prompt $b$ \\
$\eta_{b,a}$, $\beta_b$ & energetic saddle of augmentation $a$; its exponential rate for prompt $b$ \\
$\mu_\alpha, \sigma_\alpha, \mu_\beta, \sigma_\beta, \rho$ & means, spreads and correlation of $(\alpha_b, \log\beta_b)$ across prompts, Eq.~(\ref{eq:ansatz}) \\
$q_b = \langle p_{b,a}\rangle_a$ & single-attempt jailbreak rate of prompt $b$ \\
$E_b = -\log q_b$; $\mu_E$, $\sigma_E$ & effective barrier, Eq.~(\ref{eq:def-E}); its mean and spread, Eq.~(\ref{eq:sigE}) \\
$Q_b(M)$, $R_b(M)$ & survival of one augmentation with $M$ completions; $R_b = -\log Q_b$, Eq.~(\ref{eq:Rb}) \\
$\ASRNM$, $\AFRNM$ & attack success and failure rate with $N$ augmentations and $M$ completions, Eq.~(\ref{eq:ASRNM}) \\
$z = NM e^{-\mu_E}$, $f_\gamma$ & scaling variable and scaling function, Eqs.~(\ref{eq:scaling-form})--(\ref{eq:f-def}) \\
$\nu_{\rm eff}$ & local exponent $-\dd\log \AFRNM/\dd\log N$, Eq.~(\ref{eq:effective-exponent}) \\
$P$, $\Nres = 1/\min_b q_b$ & number of prompts; resolution budget of the dataset \\
$c_b$ & second shape parameter of the Beta soft wall, $p_{b,a} \sim B(\beta_b T, c_b)$ \\
$G_b$, $H_b$ & augmentations drawn and jailbroken in the $10^4$-augmentation arm \\
$\epsilon$ & judge false-positive rate per completion \\
\bottomrule
\end{tabular}}
\end{table}

\end{document}